\documentclass{article}

\usepackage[preprint,nonatbib]{neurips_2022}

\usepackage[utf8]{inputenc} % allow utf-8 input
\usepackage[T1]{fontenc}    % use 8-bit T1 fonts
\usepackage{hyperref}       % hyperlinks
\usepackage{url}            % simple URL typesetting
\usepackage{booktabs}       % professional-quality tables
\usepackage{amsmath}
\usepackage{amsfonts}       % blackboard math symbols
\usepackage{nicefrac}       % compact symbols for 1/2, etc.
\usepackage{microtype}      % microtypography
\usepackage{xcolor}         % colors
\usepackage{multirow}
\usepackage{graphicx}
\usepackage{subfigure}
\usepackage{hyperref}
\usepackage{wrapfig}
\usepackage{algorithmic}
\usepackage[ruled]{algorithm2e}

\SetKwInput{KwInput}{Input}
\SetKwInput{KwOutput}{Output}
\DeclareMathOperator*{\argmin}{arg\,min}

\title{Enhance the Visual Representation via Discrete Adversarial Training}

\author{%
  Xiaofeng Mao$^{\dagger}$ \quad Yuefeng Chen$^{\dagger}$ \quad Ranjie Duan$^{\dagger}$ \quad Yao Zhu$^{\ddagger}$ \quad Gege Qi$^{\dagger}$\\ 
  \textbf{\quad Shaokai Ye$^{\S}$ \quad Xiaodan Li$^{\dagger}$ \quad Rong Zhang$^{\dagger}$ \quad Hui Xue$^{\dagger}$}\\
  $^{\dagger}$Alibaba Group, $^{\ddagger}$Zhejiang University, $^{\S}$EPFL\\
  \texttt{\{mxf164419,yuefeng.chenyf,ranjie.drj\}@alibaba-inc.com} \\
}

\begin{document}

\maketitle

\begin{abstract}
  
Adversarial Training (AT), which is commonly accepted as one of the most effective approaches defending against adversarial examples, can largely harm the standard performance, thus has limited usefulness on industrial-scale production and applications. Surprisingly, this phenomenon is totally opposite in Natural Language Processing (NLP) task, where AT can even benefit for generalization. We notice the merit of AT in NLP tasks could derive from the discrete and symbolic input space. For borrowing the advantage from NLP-style AT, we propose Discrete Adversarial Training (DAT). DAT leverages VQGAN to reform the image data to discrete text-like inputs, i.e. visual words. Then it minimizes the maximal risk on such discrete images with symbolic adversarial perturbations. We further give an explanation from the perspective of distribution to demonstrate the effectiveness of DAT. As a plug-and-play technique for enhancing the visual representation, DAT achieves significant improvement on multiple tasks including image classification, object detection and self-supervised learning. Especially, the model pre-trained with Masked Auto-Encoding (MAE) and fine-tuned by our DAT without extra data can get \textbf{31.40} mCE on ImageNet-C and \textbf{32.77\%} top-1 accuracy on Stylized-ImageNet, building the new state-of-the-art. The code will be available at \url{https://github.com/alibaba/easyrobust}.

\end{abstract}

\section{Introduction}

Nowadays, Deep Neural Networks (DNNs) has achieved excellent performance surpassing humans in most computer vision tasks. Although remarkable progress has been made, the success of DNNs is actually a false sense when i.i.d hypothesis is not satisfied in wild. Researchers have shown that deep models fail in most circumstances including adversarial perturbations~\cite{Szegedy2014IntriguingPO}, common corruptions~\cite{Hendrycks2019BenchmarkingNN}, colors or textures changing~\cite{geirhos2018imagenet, shamsabadi2020colorfool}, etc. There is still a long way to make DNNs closer to the robust human perception. 

A possible way towards robust machine perception can be Adversarial Training (AT)~\cite{madry2018towards}, which automatically finds failure input cases of DNNs and augment online with these cases for fixing "bugs". 
With online augmentation of adversarial examples, AT greatly enhances the adversarial robustness, and helps for learning perceptually-aligned representations~\cite{engstrom2019learning} with good interpretability~\cite{ross2018improving,ilyas2019adversarial} and transferability~\cite{salman2020adversarially}. However, AT is double-edged, which meanwhile degrades the standard performance caused by problematic regularization~\cite{wen2020towards}. Such problematic regularization makes the decision boundaries over-smoothed and enlarges indecisive regions. 

Surprisingly, previous works~\cite{zhu2019freelb,ivgi2021achieving} observe a strange phenomenon that AT behaves conversely in Natural Language Processing (NLP) tasks. By automatically finding adversarial textual inputs, AT will not hurt the accuracy and even benefit for both generalization and robustness of language models. This phenomenon motivates us considering whether the merit of NLP-style AT can be transferred to vision tasks. 
% To validate it, we assume such merit is derived from the unique data organizing form of NLP tasks. 
We notice such merit could derive from the unique data organizing form of language models. To be specific, an adversarial image perturbed in continuous pixel space actually differs with the truly "hard" examples appeared in real world. Contrarily, text space is discrete and symbolic, where adversarial text is practically existing when a typo is made by humans. Learning on such adversarial text will obviously improve the generalization on other more texts with confusing typos.
% Since symbolic perturbations make the adversarial texts closer to natural distribution
% Since adversarial text with symbolic perturbations Closer to natural distribution
% the perturbations on text are discrete and symbolic. Compared with adversarial image which actually differs with the truly "hard" examples appeared in real world, the textual perturbations are usually meaningful and practically existing in real-world when a typo is made by humans. % Although such textual perturbations is not the worst or along the steepest gradient direction, they are
% % Different from adversarial image perturbations, such textual perturbations is not the worst or along the steepest gradient direction. However, it is much meaningful than adversarial images as  closer to real-world data distribution, where a typo is always happens in  , the adversarial perturbations on text are  on images which
% % Compared with images, the perturbations on text are symbolic, which causes that text adversarial examples are sparse and exhaustible in input space. It will be much easier to construct ideal hyperplanes to classify these sparse distributed samples. 
Therefore, we borrow the symbolic nature of languages, and apply it on CV tasks by discretizing continuous images into a more meaningful symbolic space.
% Based on above supposition, we consider to mimick the standard input of language models by discretizing a continuous image into visual words.
Afterwards, AT is conducted for minimizing the maximal risk on such text-like inputs with symbolic adversarial perturbations.

\begin{figure}
  \centering
  \includegraphics[width=1.\linewidth]{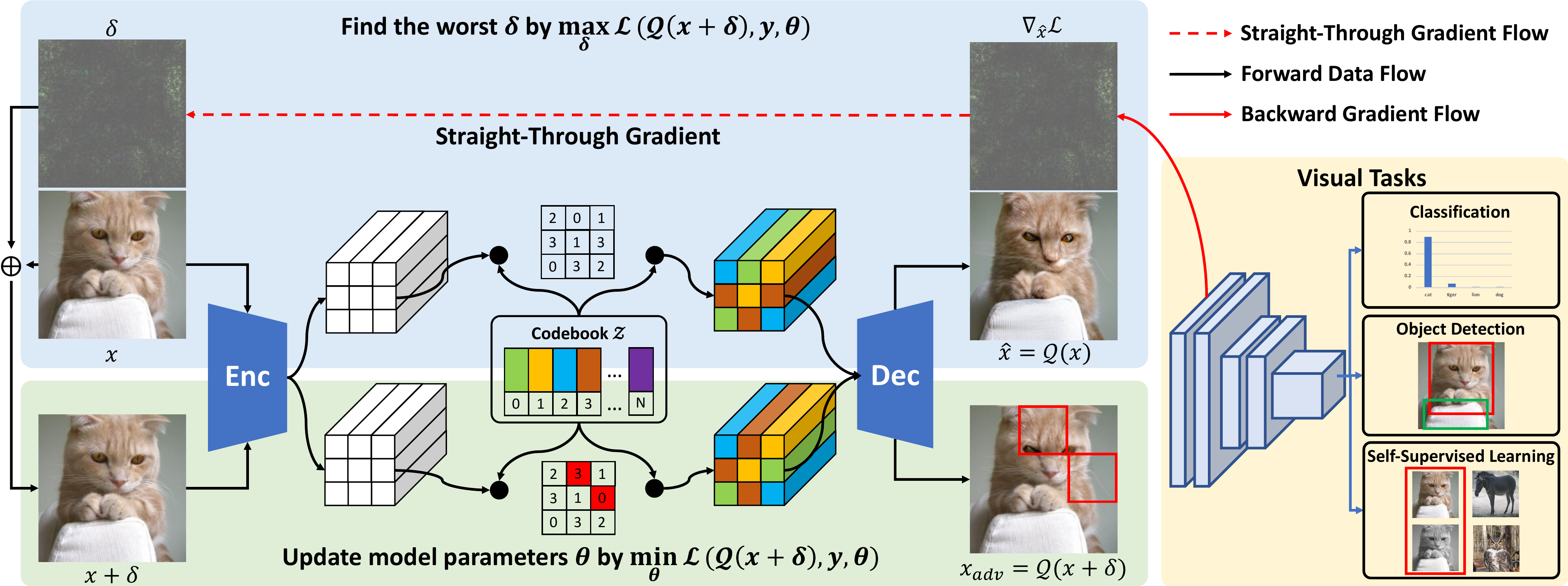}
  \caption{The overall pipeline of Discrete Adversarial Training (DAT). }
  \label{fig:main}
  \vspace{-1.5em}
\end{figure}

In this paper, we propose Discrete Adversarial Training (DAT), a new type of adversarial training which aims to improve both robustness and generalization of vision models. DAT leverages VQGAN~\cite{esser2021taming} to learn a vocabulary of visual words, also known as image codebook. For a continuous image input, each encoded patch embedding is replaced with its closest visual word in the codebook, and represented as a corresponding index. Then the image is transformed to a sequence of symbolic indices similar with language input. For generating adversarial examples based on such symbolic sequence, direct use of optimization methods in NLP like combinatorial optimization~\cite{jin2020bert, zang2020word, alzantot2018generating} or synonym substitutions~\cite{jin2020bert, alzantot2018generating} can be challenging. The reason lies in: 1) the large search space of images and 2) the non-existence of synonym in visual codebook. To make it more efficient, DAT adopts a gradient-based method which assumes the backward adversarial gradient goes straight-through the complex discretization process, thus gradients on discretized image can be copied to original input. Then it use one-step search along the direction of estimated gradient such that the discrete representation will altered adversarially during the discretization process, resulting a discrete adversarial examples. Finally the discrete adversarial examples is fed into models for training. Different from AT which always adds $l_{p}$ bound on augmented adversarial examples, DAT affects the  discretization process to produce diverse adversarial inputs beyond $l_{p}$ bound for training. We show in ablation experiment that DAT not only enhances the robustness on $l_{p}$ bounded attacks, but also is partly beneficial in defense of unrestricted semantic attacks~\cite{duan2021advdrop,zhao2020towards}. The overall pipeline of discrete adversarial training is shown in Figure~\ref{fig:main}. 

We further give an analysis to explain the effectiveness of our DAT from the perspective of distribution. By comparing the distributional difference of training examples in AT and DAT with clean images, we find the discrete adversarial examples in our DAT are much closer to the clean distribution. Such ability of generating "in-distribution" adversarial examples makes DAT can improve the visual representation learning on multiple vision models and tasks, with no sacrificing of clean accuracy.

% There are three natural advantages for discrete adversarial training. 
% 1) Due to the sparse input space, DAT adds gentler regularization on models, which alleviates the standard performance degradation. 
% Meanwhile, with sufficient training on discrete adversarial examples, the robustness can be improved significantly. 
% This finding is also consistent with what we observed in NLP. 
% 2) Different from AT which always adds $l_{p}$ bound on augmented adversarial examples, DAT perturbs the discrete latent instead of pixels, and thus can produce diverse adversarial inputs beyond $l_{p}$ bound for training. We show DAT not only enhances the robustness on $l_{p}$ bounded attacks, but also is partly beneficial in defense of unrestricted semantic attacks~\cite{duan2021advdrop,zhao2020towards}. 
% 3) As a plug-and-play technique, DAT can improve the quality of visual representation on multiple vision tasks. 

Our contributions are summarized below:
\begin{itemize}
\item To the best of our knowledge, we appear to be the first to transfer the merit of NLP-style adversarial training to vision models, for improving robustness and generalization simultaneously.

\item We propose Discrete Adversarial Training (DAT), where images are presented as discrete visual words, and the model is training on example which has the adversarially altered discrete visual representation. 

\item  DAT achieves significant improvement on multiple tasks including image classification, object detection and self-supervised learning. Especially, it establishes a new state-of-the-art for robust image classification. By combining MAE~\cite{he2021masked} pre-training and DAT fine-tuning, our ViT-Huge~\cite{dosovitskiy2020image} achieves 31.40 mCE on ImageNet-C~\cite{Hendrycks2019BenchmarkingNN} and 32.77\% top-1 accuracy on Stylized-ImageNet~\cite{geirhos2018imagenet}.

\end{itemize}

\section{Related Work}
 
\paragraph{Adversarial Training} Adversarial Training (AT) ~\cite{madry2018towards} is first proposed to improve robustness by training models with adversarial examples. As one of the most effective defense, existing works~\cite{tsipras2018robustness, zhang2019theoretically} have suggested a trade-off between adversarial and clean accuracy in AT. Despite great efforts~\cite{raghunathan2020understanding, rade2021reducing,lamb2019interpolated} have been made for mitigating this trade-off, the bad generalization of AT still cannot be fully remedied till now. In opposite way, some other works~\cite{xie2020adversarial,mei2021fast} use AT to improve the clean accuracy rather than adversarial robustness. The most close work to ours is the AdvProp~\cite{xie2020adversarial}, which splits batch norms to prevent the mixed statistics of clean and adversarial examples, thus learns better adversarial feature for generalization. Pyramid AT~\cite{herrmann2021pyramid} makes AT specific to ViTs by crafting pyramid perturbation with Dropout enabling, yielding imporved performance. However, these methods are only applicable under specific models or tasks. VILLA~\cite{gan2020large} is also a representation enhancement technique using AT. But contrary to us, it applies vision-style AT only for vision-and-language representation learning. AGAT~\cite{gokhale2021attribute} is another kind of AT beyond pixel space, which perturbs images along attributes, however it has strict requirement of attributes annotation.

\vspace{-1em}

\paragraph{Adversarial Augmentation}
By borrowing the idea of AT, some previous works~\cite{zhang2019adversarial, wang2021augmax, calian2021defending} propose to search augmentations adversarially for improving the hardness of training examples. Adversarial AutoAugment~\cite{zhang2019adversarial} uses augmentation policy network to produce hard augmentation policies on a pre-defined policy space. AugMax~\cite{wang2021augmax} mixes multiple randomly sampled augmentation operators like AugMix~\cite{hendrycks2019augmix}, by using adversarially learned mixing factors. AdA~\cite{calian2021defending} optimizes the parameters of image-to-image models to generate adversarially corrupted augmented images. MaxUp~\cite{gong2021maxup} uses the worst augmented data of each data point in a set of random perturbations or transforms for training. However, these methods create adversarial inputs indirectly and rely on pre-defined augmentations or translation models. They cannot perturb the images locally. Instead, our DAT directly modifies the image with no need of pre-defined transforms, and can craft local perturbations on images, which is more elaborate.

\vspace{-1em}

\paragraph{Discrete Visual Representation Learning}
Early technique~\cite{sivic2003video, csurka2004visual} of Bag-Of-Visual-Words (BOVW) model has shown great power of discrete representation in visual understanding. VQ-VAE~\cite{van2017neural} uses DNNs to learn neural discrete representations, also known as visual codebook, by generative modeling the image distribution. Recently, the idea of discrete representations learning has been widely emerged in many vision tasks. In most Masked Image Modeling (MIM) methods~\cite{bao2021beit, zhou2021ibot}, visual codebook is needed for BERT-like self-supervised pretraining. For image classification, discrete representations strengthen the robustness by preserving the global structure of an object and ignoring local details~\cite{mao2021discrete}. For image synthesis, adversarial and perceptual objective can be added to VQ-VAE for learning perceptually-aligned visual codebook~\cite{esser2021taming}. With growing power of generative models, now a well trained VQGAN can produce vivid images with 0.58 FID~\cite{rombach2021high}. In this work, we use VQGAN for the discretization process in our DAT. As a powerful generative model, VQGAN can greatly reduce the information loss in reconstruction process and ensure the high-quality of the generated discrete adversarial examples.

\section{Method}

\subsection{Traditional Adversarial Training}
We take the typical image classification task as an example to show the formulation of Adversarial Training (AT). Consider the training image and label set $\mathcal{D} = \{x_{i}, y_{i}\}_{i=1}^{n}$, and a classifier $F$ with learnable parameters $\theta$, the classification objective is always a cross-entropy loss $\mathcal{L}(x, y, \theta)$. Adversarial Training (AT) finds the optimal $\theta$ by solving a minimax optimization problem:
% \begin{equation}
%     \min_{\theta} \max_{\delta} \mathcal{L}(x+\delta, y, \theta) \ \ s.t. \|\delta \|_{p} < \epsilon,
% \label{eq:at}
% \end{equation}
\begin{equation}
    \min_{\theta} \mathbb{E}_{(x, y) \sim \mathcal{D}} \left[ \max_{\delta} \mathcal{L}(x+\delta, y, \theta) \right] \ \ s.t. \|\delta \|_{p} < \epsilon,
\label{eq:at}
\end{equation}
where the inner optimization finds the perturbations $\delta$ on per-pixel values for maximizing the loss, and the outer minimization update $\theta$ to improve the worst-case performance of the network w.r.t. the perturbation. $\|\cdot\|_{p}$ constraints the $p$-norm of $\delta$ to a small value $\epsilon$. A problem is that AT finds the failure case i.e. $x+\delta$ in continuous pixel space. However, human does not create or recognize images from complex pixel values, but from discrete semantic concepts. Although adversarial examples can successfully fool the models, they are still different from the real "hard" examples appeared in practice.

\subsection{Discrete Adversarial Training}
\subsubsection{Image Discretization by Visual Codebook}
\label{sec:3.2.1}
For discrete adversarial training, it is desirable first to learn an expressive visual codebook and represent the training image set in discrete space. We utilize VQGAN~\cite{esser2021taming} for image discretization. 
More precisely, consider a continuous image $x \in \mathbb{R}^{H\times W\times 3} $, VQGAN learns an encoder $\mathsf{Enc}_\phi\left(\cdot\right)$, decoder $\mathsf{Dec}_\psi\left(\cdot\right)$ and quantization $\mathsf{q}_\mathcal{Z}\left(\cdot\right)$. 
$\mathsf{Enc}_\phi$ is a convolutional model which maps $x$ to intermediate latent vectors $v = \mathsf{Enc}_\phi(x) \in \mathbb{R}^{ (h\times w) \times d}$, where $h, w$ is the height, width of the intermediate feature map and $d$ is the latent dimension. 
Subsequently, $\mathsf{q}_\mathcal{Z}(\cdot)$ learn a codebook $\mathcal{Z} = \{z_{k}|z_{k}\in  \mathbb{R}^{d} \}_{k=1}^{K}$, such that each latent vector $v_{ij}\in \mathbb{R}^{d}$ can be quantized onto its closest codebook entry $z_{k}$ as:
\begin{equation}
    v_{\mathsf{q}} = \mathsf{q}_{\mathcal{Z}}(v) := \left(\argmin_{z_{k}\in \mathcal{Z}} \left \| v_{ij} - z_{k} \right \| \right) \in \mathbb{R}^{h\times w\times d}, 
\label{eq: 3.2.1}
\end{equation} where $i,j$ present each location in feature map. Then the decoder $\mathsf{Dec}_\psi\left(\cdot\right)$ outputs the reconstruction image $\hat{x}$ from the quantized vectors $v_{\mathsf{q}}$ by:
\begin{equation}
    \hat{x} =\mathsf{Dec}_\psi\left( v_{\mathsf{q}}\right).
\end{equation}

VQGAN is trained by minimizing the reconstructed difference between $\hat{x}$ and $x$. More details can be referred to Appendix A. So far, given a continuous image $x$, we can get its corresponding discrete reconstruction $\hat{x}$. For simplicity, we use $\mathcal{Q}$ to stand for the above image discretization process, and then we have $\hat{x} = \mathcal{Q}(x)$. 

\subsubsection{Discrete Adversarial Training}
% \begin{assumption}
% Given continuous image set $X=\{x_{i}\}_{i=1}^{n}$, we assume that a perfect $\mathcal{Q}$ which reconstructs $\hat{x} \in \mathcal{R}(x)$ satisfies: $\hat{x} =\mathcal{Q}(x)$ \textit{i.i.f} $r \to 0$, where $\mathcal{R}(x)$ is the $r$-ball centered at data point $x$.
% \label{ap:1}
% \end{assumption}

%%%%%%%%%%%%%%%%%%%%%%%%%%%%%%%%%%%%%%%%%%%%%
%%%%%%%%%%%%%%%%%%%%%%%%%%%%%%%%%%%%%%%%%%%%%

Based on the definition in Sec~\ref{sec:3.2.1}, we can generate discrete adversarial examples in inner maximization step of AT. By slightly modifying the Eq~\ref{eq:at}, the objective of DAT is formulated as:
% \begin{equation}
%     \min_{\theta} \max_{\delta} \mathcal{L}(\mathcal{Q}(x+\delta), y, \theta) \ \ s.t. \|\delta \|_{p} < \epsilon.
% \label{eq:dat}
% \end{equation}
\begin{equation}
    \min_{\theta} \mathbb{E}_{(x, y)\sim \mathcal{D}} \left[ \max_{\delta} \mathcal{L}(\mathcal{Q}(x+\delta), y, \theta) \right],
\label{eq:dat}
\end{equation}
where $\mathcal{Q}$ transforms the continuous pixel space to discrete input space. We delete the constraint term since there is no need to bound the per-pixel values of $\delta$. Suppose that $\mathcal{Q}$ is an ideal discretizer with no information loss in discretization process. The problem lies how to find the worst $\delta$ for maximizing the classification loss. Similar with traditional AT, we can use gradient-based methods to approximate $\delta$ by:
%Similar with traditional AT, we uses gradient-based methods to approximate $\delta$ by:
%This could be seen as a relatively inaccurate approximation of the inner maximization for `∞ perturbations, and has the following closed form (Goodfellow et al., 2014):
\begin{equation}
    \delta \simeq \alpha \nabla_{x}\mathcal{L}(\mathcal{Q}(x), y, \theta),
\label{eq:gradient}
\end{equation}
where $\alpha$ determines the magnitude of the perturbations along the gradient direction. We set $\alpha=0.1$ by default in DAT. To expand $\nabla_{x}\mathcal{L}(\mathcal{Q}(x), y, \theta)$ by chain rule, we have four partial derivative terms as follows:
\begin{equation}
    \nabla_{x}\mathcal{L}(\mathcal{Q}(x), y, \theta) = \frac{\partial \mathcal{L}}{\partial \hat{x}}\cdot\frac{\partial \hat{x}}{\partial v_{\mathsf{q}}}\cdot\frac{\partial v_{\mathsf{q}}}{\partial v}\cdot\frac{\partial v}{\partial x}
\label{eq:expand}
\end{equation}
Through analysing the feasibility of each term, we find only $\frac{\partial v_{\mathsf{q}}}{\partial v}$ is hard to solve as the non-differentiable nature of Eq~\ref{eq: 3.2.1}. Fortunately, as proposed in previous work, a straight-through gradient estimator~\cite{bengio2013estimating, yin2019understanding} can be used by copying the gradients from $v_{\mathsf{q}}$ to $v$. By replacing $\frac{\partial \hat{x}}{\partial v_{\mathsf{q}}}\cdot\frac{\partial v_{\mathsf{q}}}{\partial v}$ with $\frac{\partial \hat{x}}{\partial v}$, we can simplify the Eq~\ref{eq:expand} to $\nabla_{x}\mathcal{L}(\mathcal{Q}(x), y, \theta) = \frac{\partial \mathcal{L}}{\partial \hat{x}}\cdot\frac{\partial \hat{x}}{\partial v}\cdot\frac{\partial v}{\partial x}$, which has derivative everywhere. Although the solution seems workable theoretically, the huge cost makes it impractical on large-scale vision tasks. The bottleneck mainly lies on that $\frac{\partial \hat{x}}{\partial v}$ and $\frac{\partial v}{\partial x}$ require the adversarial gradients backward through $\mathsf{Enc}$ and $\mathsf{Dec}$. 
Actually, a generator capable of producing high-quality images always has a large amount of parameters. 
Compared with original adversarial training which only needs $F$ for gradient calculation, it requires more than tripled GPU memory and computation cost.

% %%%%%%%%%%%%%%%%%%%%%%%%%%%%%%%%%%%%%%%%%%%%%
% \begin{wrapfigure}{r}{0.38\textwidth}
% \vspace{-0.8mm}
% \centering
% \includegraphics[width=0.38\textwidth]{paper_fig/ae_vis.pdf}
% \caption{Visiualization of Adversarial Examples (AEs) crafted in pixel space and discrete latent space. The attack success rate is obtained by attacking ResNet50 on ImageNet validation set.}
% \vspace{-4mm}
% \label{fig:ae_vis}
% \end{wrapfigure}
% %%%%%%%%%%%%%%%%%%%%%%%%%%%%%%%%%%%%%%%%%%%%%

To solve this problem, we propose an efficient alternative solution. Since $\hat{x} \simeq x$ is empirically observed for an ideal discretizer $\mathcal{Q}$, we can also use a straight-through estimator between $\hat{x}$ and $x$, which is given by
\begin{equation}
    \nabla_{x}\mathcal{L}(\mathcal{Q}(x), y, \theta) = \frac{\partial \mathcal{L}}{\partial \hat{x}}\cdot\frac{\partial \hat{x}}{\partial x} \simeq \frac{\partial \mathcal{L}}{\partial \hat{x}}.
\label{eq:final}
\end{equation}

Finally, we can solve the worst $\delta$ by $\nabla_{\hat{x}}\mathcal{L}(\hat{x}, y, \theta)$. By this way, the computation cost of DAT has been largely reduced. Compared with original adversarial training, it only has extra computation cost on VQGAN forward, which is relatively controllable. 

For clarity, let us restate the pipeline of our DAT. For each training image $x$, DAT first use VQGAN to get discrete reconstruction $\hat{x}$. By feeding $\hat{x}$ to classifier $F$, a worst-case perturbation $\delta$ can be estimated by computing the gradient of $\hat{x}$ towards maximizing the classification loss. The perturbed image thus can be created by adding $\delta$ on original $x$. Finally, $x+\delta$ is discretized by VQGAN again and acts as the adversarial input, on which $F$ is trained by minimizing the classificaiton loss. The details of our DAT is summarized in Algorithm~\ref{alg:dat}.

%%%%%%%%%%%%%%%%%%%%%%%%%%%%%%%%%%%%%%%%%%%%%
\begin{wrapfigure}{r}{0.5\textwidth}
\vspace{-6mm}
\centering
\includegraphics[width=0.5\textwidth]{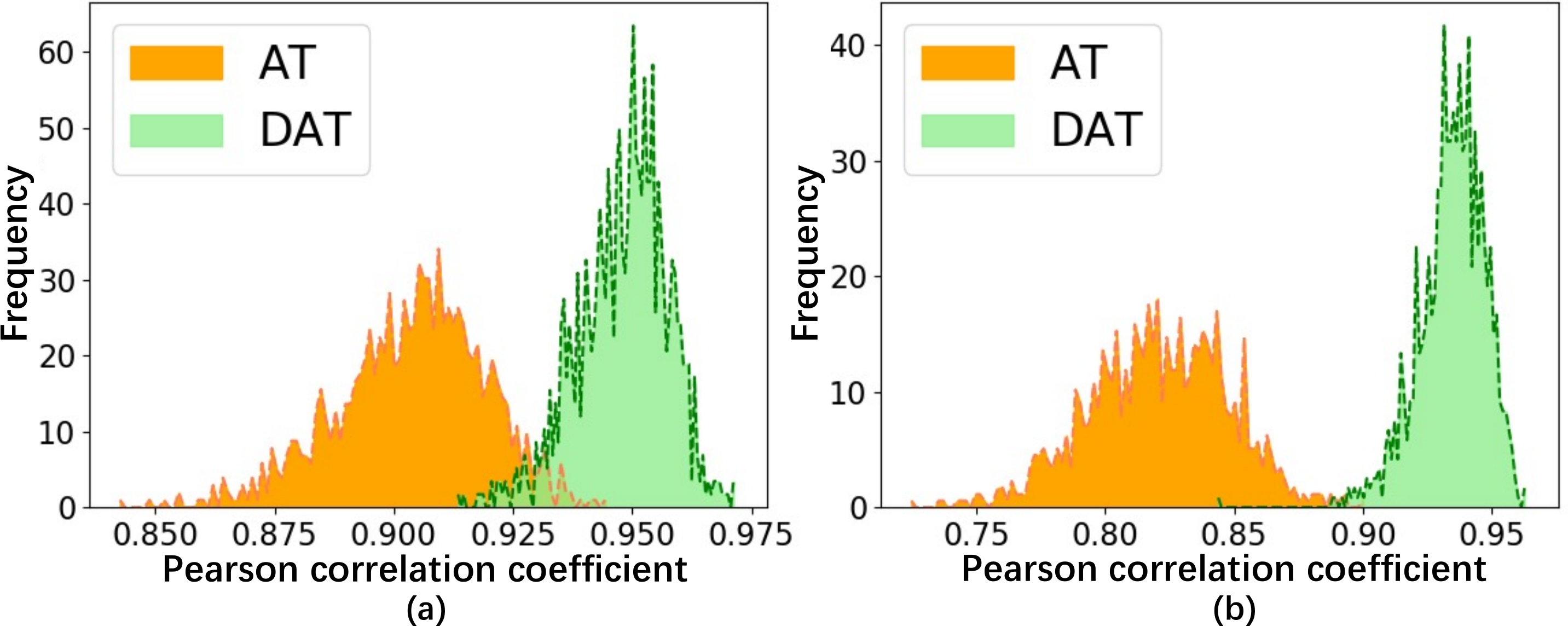}
\vspace{-6mm}
\caption{The frequency histogram of the Pearson correlation coefficient (PCC) between BN statistics on clean and adversarial images. Larger PCC value means smaller distributional difference with clean images. (a), (b) present the difference on mean and variance statistics respectively. }
\vspace{-3mm}
\label{fig:ae_vis}
\end{wrapfigure}
\begin{figure}
\begin{algorithm}[H]
\caption{Pseudo code of DAT}
\KwInput{Classifier $F$; Pre-trained discretizer $\mathcal{Q}$; A sampled mini-batch of clean images $x$ with labels $y$;  attack magnitude $\alpha$.}
\KwOutput{Learned network parameter $\theta$ of $F$}
\begin{algorithmic}[1]
\STATE Fix the network parameters of $\mathcal{Q}$
\FOR{each training steps}
\STATE $\hat{x} \leftarrow \mathcal{Q}(x)$\hfill\COMMENT{Get the discrete reconstruction $\hat{x}$}
\STATE $\delta \leftarrow \alpha \nabla_{\hat{x}}\mathcal{L}(\hat{x},y,\theta)$ \hfill\COMMENT{Estimate the adversarial perturbations}

\STATE $x_{adv} \leftarrow \mathcal{Q}(x+\delta)$ \hfill\COMMENT{Generate discrete adversarial examples}
\STATE Minimize the classification loss w.r.t. network parameter
\\ 
$\argmin_{\theta}\mathcal{L}(x_{adv}, y, \theta)$
\ENDFOR
\end{algorithmic}
\label{alg:dat}
\end{algorithm}
\vspace{-2em}
\end{figure}

\paragraph{Explaining the Effectiveness of DAT from the Perspective of Distribution} 
% It is still not clear why DAT can improve the robustness and generalization of visual representations without sacrificing clean accuracy, and which good property of DAT explains its superiority compared with AT. 
We give an empirical analysis to explain why DAT can improve the robustness and generalization without sacrificing clean accuracy. Previous work~\cite{xie2020adversarial} has pointed out that the underlying distributions of adversarial examples are different from clean images. Training on both clean and adversarial images will force the Batch Normalization (BN)~\cite{ioffe2015batch} to estimate an inaccurate mixture statistics of feature distribution, and thus impact the standard performance. We study this effect by sampling 1000 mini-batches in ImageNet validation set, and generate corresponding adversarial images for AT and our DAT. For each batch we calculate the mean and variance statistics of last BN of ResNet50. Then the Pearson correlation coefficient (PCC) between the statistics on clean and adversarial examples is computed for showing the distributional difference. Figure~\ref{fig:ae_vis} shows the frequency histogram of the distributional difference on 1000 mini-batches. For training samples of DAT, the peak of the histogram is at 0.95, which is greater than AT. It suggests that DAT generates discrete adversarial examples much closer with the clean distributions. Therefore, training on these examples will reduce the shift of clean distribution in AT, yielding both the robustness and generalization improvement.

\section{Experiments}
To demonstrate the versatility of our method, we experiment Discrete Adversarial Training (DAT) on multiple tasks including image classification, object detection and self-supervised learning.

\subsection{Image Classification}
\label{sec:4.1}
\paragraph{Implementation} We implement DAT on two representative architectures: ResNet50~\cite{he2016deep} and ViTs~\cite{dosovitskiy2020image}. For ResNet50, we first experiment DAT with vanilla training recipes using "robustness" library \footnote{https://github.com/MadryLab/robustness}. Then we combine DAT with other orthogonal robust training techniques: DeepAugment~\cite{hendrycks2021many} and AugMix~\cite{hendrycks2019augmix}. Only cross entropy loss is used for generating discrete adversarial examples. The JSD loss in AugMix is optimized merely on clean samples. For ViTs, we adopt ViT-B/16 as baseline models, which is trained by the recipes in AugReg~\cite{steiner2021train}. Besides, we use DAT to conduct supervised finetuning on downstream ImageNet classification task based on a self-supervised ViT-Huge pretrained by MAE~\cite{he2021masked}. By default, we refer ViT to ViT-B/16 in all tables and figures. 

\paragraph{Benchmarks} The trained model is evaluated in three aspects: 1) in-distribution performance on ImageNet-Validation set; 2) adversarial robustness on white-box FGSM~\cite{goodfellow2014explaining} and transfer-based black-box attack dataset DamageNet~\cite{chen2020universal}; 3) out-of-distribution robustness on ImageNet(IN)-A, IN-C, IN-V2, IN-R, IN-Sketch and Stylized IN. Each of them represents a type of out-of-distribution scenario where the classifier is prone to make mistakes. IN-A~\cite{hendrycks2021natural} places the ImageNet objects in hard contexts; IN-C~\cite{hendrycks2018benchmarking} applies a series of noise, blur, digital and weather corruptions; IN-R~\cite{hendrycks2021many} collects online images with artificial creation, e.g., cartoons, graphics, video game renditions, etc; IN-Sketch~\cite{wang2019learning} contains images described by sketches; Stylized IN~\cite{geirhos2018imagenet} destroys the texture but maintains the shape feature by conducting style transfer on ImageNet images. Except for IN-C which is measured by mCE, we report the top-1 accuracy on all above datasets. 

\paragraph{Results} We report all results in Table~\ref{tab:main_result_cls}. For fair comparison, we add DAT on base methods without modifying the original training hyper-parameters, such that improvement is entirely attributed to the DAT. For ResNet50, DAT achieves significant improvement on both clean accuracy, adversarial and out-of-distribution robustness. The improvement seems greater when combining DAT with DeepAugment and AugMix. For ViTs, we find DAT is compatible with other complex augmentations such as MixUp~\cite{zhang2018mixup}, CutMix~\cite{yun2019cutmix} or RandAugment~\cite{cubuk2020randaugment} used in AugReg, yielding greater improvement. Compared with plain ViT, AugReg-ViT and ViT with discrete representation called DrViT~\cite{mao2021discrete}, DAT with AugReg achieves better performance. The best result is from ViT-Huge pretrained by MAE~\cite{he2021masked} and finetuned by DAT, which suggests DAT is also effective in downstream fine-tuning tasks.

Additionally, we also compare DAT with other robust training strategies in Table~\ref{tab:strategy_compare}. AugReg-ViT is adopted as the baseline model. Most strategies, e.g., AdvProp~\cite{xie2020adversarial} and Debiased~\cite{li2020shape} are proposed for ResNet with auxiliary BatchNorm. We show these methods cannot work properly on ViTs with only LayerNorm. Compared with Pyramid AT~\cite{herrmann2021pyramid}, our DAT has lower clean accuracy but yields stronger robustness. More results of strategies comparison on ResNet50 refer to Appendix B. 

%%%%%%%%%%%%%%%%%%%%%%%%%%%%%%%%%%%%%%%%%%%%%
\begin{table*}\centering
\scriptsize
\begin{tabular}{l|c|cc|ccc|ccc}
\toprule
\multirow{ 2}{*}{Methods} & \multirow{ 2}{*}{ImageNet} & \multicolumn{2}{c|}{Adversarial Robustness} & \multicolumn{6}{c}{Out of Distribution Robustness} \\
 & & FGSM & DamageNet & A & C$\downarrow$ & V2 & R & Sketch & Stylized \\
\midrule
 ResNet50~\cite{he2016deep} & 76.13 & 12.19 & 5.94 & 0.0 & 76.70 & 63.20 & 36.17 & 24.09 & 7.38 \\
 + \textbf{DAT (Ours)}& \textbf{76.52} & \textbf{30.66} &  \textbf{14.42} & \textbf{4.38} & \textbf{74.16} & \textbf{65.02} & \textbf{41.90} & \textbf{27.27} & \textbf{10.8} \\
 \midrule
 DeepAugment~\cite{hendrycks2021many} & 76.66 & 21.61 & 11.94 & 3.46 & 60.37 & \textbf{65.24} & 42.17 & 29.50 & 14.68 \\
 + Augmix~\cite{hendrycks2019augmix} & 75.82 & 27.05 & 19.60 & 3.86 & 53.55 & 63.63 & 46.77 & 32.62 & 21.23 \\
 \textbf{+ DAT (Ours)} & \textbf{77.10} & \textbf{35.32} & \textbf{22.86} & \textbf{6.86} & \textbf{50.82} & 65.14 & \textbf{47.88} & \textbf{34.98} & \textbf{21.89} \\
\midrule
 ViT~\cite{dosovitskiy2020image} & 72.00 & 23.30 & 28.99 & 6.44 & 77.61 & 57.34 & 25.69 & 15.56 & 5.82 \\
 DrViT~\cite{mao2021discrete} & 79.48 & 45.76 & 44.91 & 17.20 & 46.22 & 68.05 & 44.77 & 34.59 & 19.38 \\
 AugReg-ViT~\cite{steiner2021train} & 79.91 & 44.32 & 45.24 & 19.03 & 54.50 & 67.90 & 39.46 & 29.16 & 16.62 \\
 \textbf{+ DAT (Ours)} & \textbf{81.46} & \textbf{51.82} & \textbf{45.70} & \textbf{30.15} & \textbf{44.65} & \textbf{70.83} & \textbf{47.34} & \textbf{34.77} & \textbf{23.13} \\
\midrule

 MAE-H~\cite{he2021masked} & 86.90 & 60.16 & 64.36 & 68.18 & 33.92 & 78.47 & 64.12 & 49.08 & 26.36 \\
 \textbf{+ DAT (Ours)} & \textbf{87.02} & \textbf{63.77} & \textbf{70.42} & \textbf{68.92} & \textbf{31.40} & \textbf{78.82} & \textbf{65.61} & \textbf{50.03} & \textbf{32.77} \\
\bottomrule
 
\end{tabular}
\vspace{-0.5em}
\caption{The results of DAT on image classification. Bold number indicates the better performance.}
\vspace{-2em}
\label{tab:main_result_cls}
\end{table*}
%%%%%%%%%%%%%%%%%%%%%%%%%%%%%%%%%%%%%%%%%%%%%

%%%%%%%%%%%%%%%%%%%%%%%%%%%%%%%%%%%%%%%%%%%%%
\begin{table*}\centering
\scriptsize
\begin{tabular}{l|c|cc|ccc|ccc}
\toprule
Training & \multirow{ 2}{*}{ImageNet} & \multicolumn{2}{c|}{Adversarial Robustness} & \multicolumn{6}{c}{Out of Distribution Robustness} \\
 Strategies &  & FGSM & DamageNet & A & C$\downarrow$ & V2 & R & Sketch & Stylized \\
 \midrule
 Normal~\cite{steiner2021train} & 79.91 & 44.32 & 45.24 & 19.03 & 54.50 & 67.90 & 39.46 & 29.16 & 16.62 \\
 Advprop~\cite{xie2020adversarial} & 79.54 & \textbf{72.38} & 45.48 & 18.53 & 51.46 & 68.74 & 43.51 & 31.68 & 19.24 \\
 Fast Advprop~\cite{mei2021fast} & 79.02 & 70.52 & 44.87 & 17.86 & 53.31 & 67.09 & 41.84 & 29.42 & 18.39 \\
 Pyramid AT~\cite{herrmann2021pyramid} & \textbf{81.68} & 50.36 & 45.53 & 23.18 & 44.95 & 70.32 & 47.30 & \textbf{36.87} & 20.02 \\
 Debiased~\cite{li2020shape} & 79.33 & 46.85 & 44.99 & 18.32 & 49.82 & 67.55 & 40.32 & 29.43 & 22.37 \\
 \textbf{DAT (Ours)} & 81.46 & 51.82 & \textbf{45.70} & \textbf{30.15} & \textbf{44.65} & \textbf{70.83} & \textbf{47.34} & 34.77 & \textbf{23.13} \\
\bottomrule
 
\end{tabular}
\vspace{-0.5em}
\caption{Comparison of DAT with other training strategies. We use AugReg-ViT as the base model.}
\label{tab:strategy_compare}
\vspace{-1.2em}
\end{table*}

%%%%%%%%%%%%%%%%%%%%%%%%%%%%%%%%%%%%%%%%%%%%%
\subsection{Self-Supervised Learning}
\paragraph{Implementation}  We experiment DAT on three self-supervised methods: SimCLR~\cite{chen2020simple}, SimSiam~\cite{chen2021exploring} and recently proposed MoCov3~\cite{chen2021empirical}. The discrete adversarial training is only conducted during pre-training stage, and the learned representation is evaluated on downstream tasks by standard pipeline~\cite{ericsson2021well}. We craft adversarial examples based on RoCL~\cite{kim2020adversarial}, which attacks the pre-training objective by maximizing the contrastive loss. For preventing the cost explosion, we pre-train 200 epochs for SimCLR, and 100 epochs for both SimSiam and MoCov3. The batch size used for SimCLR, SimSiam, MoCov3 is set as 1024, 512, 2048 respectively.

\paragraph{Benchmarks} For self-supervised learning, we adopt ImageNet-1K for both training and in-distribution testing. Beyond that, in order to give more comprehensive assessment, we build three downstream tasks to metric the transferability of the learned visual representation. Specifically, the linear evaluation reports the top-1 accuracy on five classification datasets: Flowers, CIFAR10, Caltech101, Cars and DTD. For downstream evaluation of object detection, we present mAP, AP50, AP75 on Pascal VOC2007~\cite{everingham2010pascal}. ADE20K~\cite{zhou2017scene} is used for semantic segmentation task, and both the mean intersection over union (mIoU) and accuracy are reported.

%%%%%%%%%%%%%%%%%%%%%%%%%%%%%%%%%%%%%%%%%%%%%
\begin{table*}\centering
\scriptsize
\begin{tabular}{l|c|ccccc|ccc|cc}
\toprule
& & \multicolumn{5}{c|}{Linear Evaluation} & \multicolumn{3}{c|}{VOC Object Detection} & \multicolumn{2}{c}{ADE20K} \\
 Method & ImageNet & Flowers & CIFAR10 & Caltech101 & Cars & DTD & mAP & AP50 & AP75 & mIoU & Acc. \\
\midrule
 MoCov3 & 68.63 & 91.54 & 93.40 & 90.38 & 49.01 & 73.03 & 50.42 & 80.53 & 53.93 & 0.3508 & 75.64 \\
 \textbf{+ DAT} & \textbf{69.60} & \textbf{93.15} & \textbf{95.16} & \textbf{91.42} & \textbf{53.09} & \textbf{73.55} & \textbf{51.92} & \textbf{80.97} & \textbf{56.04} & \textbf{0.3585} & \textbf{76.33} \\
 \midrule
 SimCLR & 64.89 & 89.28 & 88.47 & 83.20 & 38.84 & \textbf{73.14} & 48.50 & 78.75 & 51.35 & 0.3396 & \textbf{75.61} \\
 \textbf{+ DAT} & \textbf{65.47} & \textbf{90.14}  & \textbf{89.97}  & \textbf{85.09} & \textbf{39.42} & 72.93 & \textbf{48.83} & \textbf{79.27} & \textbf{51.95} & \textbf{0.3412} & 75.46 \\
 \midrule
 SimSiam & 68.16 & \textbf{87.67} & 89.45 & 85.94 & 34.95 & 71.70 & 48.92 & 77.22 & 52.69 & \textbf{0.2212} & 68.57 \\
 \textbf{+ DAT} & \textbf{68.41} & 86.93 & \textbf{91.70} & \textbf{87.03} & \textbf{35.10} & \textbf{73.03} & \textbf{51.73} & \textbf{79.69} & \textbf{55.74} & 0.2203 & \textbf{68.92} \\
 \bottomrule
\end{tabular}
\vspace{-0.5em}
\caption{The results of DAT on self-supervised learning.}
\label{tab:main_result_ssl}
\vspace{-2.5em}
\end{table*}
%%%%%%%%%%%%%%%%%%%%%%%%%%%%%%%%%%%%%%%%%%%%%

%%%%%%%%%%%%%%%%%%%%%%%%%%%%%%%%%%%%%%%%%%%%%
\begin{wraptable}{r}{0.58\textwidth}
\scriptsize
\begin{tabular}{c|c|c|c|c}
\toprule
\multirow{ 2}{*}{Models} & Training & COCO & COCO-C &  Relative \\
   & Strategy & mAP & mAP & rPC (\%) \\
\midrule
\multirow{ 3}{*}{EffDet-Lite0-320~\cite{tan2020efficientdet}} & Normal & 26.41 & 16.11 & 61.00  \\
 &  Det-Advprop & 26.34 & 16.38 & 62.19 \\ 
 &  \textbf{DAT (Ours)} & \textbf{27.32} & \textbf{17.89} & \textbf{65.48} \\ 
\midrule
\multirow{ 3}{*}{EffDet-Lite1-384~\cite{tan2020efficientdet}} & Normal & 31.50 & 19.43 & 61.68 \\ 
 &  Det-Advprop & 31.82 & 20.21 & 63.51 \\ 
 &  \textbf{DAT (Ours)} & \textbf{32.31} & \textbf{21.32} & \textbf{65.99} \\ 
\midrule
\multirow{ 3}{*}{YOLOv3-320~\cite{redmon2018yolov3}} & Normal & 35.91 & 18.39 & 51.21 \\ 
 &  Det-Advprop & 35.73 & 19.34 & 54.13 \\ 
 &  \textbf{DAT (Ours)} & \textbf{36.02} & \textbf{20.55} & \textbf{57.05} \\ 
\midrule
\multirow{ 3}{*}{YOLOv3-416~\cite{redmon2018yolov3}} & Normal & 40.30 & 21.19 & 52.58\\ 
 &  Det-Advprop & \textbf{40.69} & 22.55 & 55.42\\
 &  \textbf{DAT (Ours)} & 40.41 & \textbf{23.38} & \textbf{57.86}  \\ 
\bottomrule
\end{tabular}
\caption{The results of DAT on object detection.}
\label{tab:main_result_det}
\vspace{-1em}
\end{wraptable}
%%%%%%%%%%%%%%%%%%%%%%%%%%%%%%%%%%%%%%%%%%%%%

\paragraph{Results} As shown in Table~\ref{tab:main_result_ssl}, DAT can enhance the learned representation on ImageNet and get 0.97\%, 0.58\% and 0.25\% improvement on MoCov3, SimCLR, SimSiam respectively. For excluding that DAT is not just over-fitting on ImageNet, we transfer the representations on downstream recognition, object detection, semantic segmentation tasks. The results suggest that DAT enhances the self-supervised representations with better transferability. In particular, MoCov3 with DAT achieves significant improvement in all downstream tasks. This conclusion echoed with previous works~\cite{salman2020adversarially}, which find adversarially robust models often perform better on transfer learning.

\subsection{Object Detection}

\paragraph{Implementation} We implement DAT on two popular detectors: EfficientDet~\cite{tan2020efficientdet} and YOLOv3~\cite{redmon2018yolov3}. Object detection generally has two sub-tasks: classification and localization.
% As figured by previous work[], attacks can adopt anyone of the objectives to fool the detectors. Therefore, adversarial training on object detection should enhance the robustness on both sub-task of classification and localization. 
Det-AdvProp~\cite{chen2021robust} proposes to select more vulnerable sub-task to generate adversarial images, and use auxiliary BN for training. Instead, DAT regards the two sub-tasks as a whole, and attacks the overall detection loss to produce adversarial images. Besides, DAT does not modify BN in detectors. As the memory and computation cost of VQGAN increased quadratic with the input resolution, DAT become unaffordable when input size is large. Therefore, we only experiment DAT on lightweight version of EfficientDet and YOLOv3 with input size smaller than 512.

\paragraph{Benchmarks} We train the detectors using the COCO 2017 object detection dataset~\cite{lin2014microsoft} and evaluate them on COCO’s validation set. COCO-C~\cite{michaelis2019benchmarking} is used for testing the robustness to natural corruptions. We report mAP on COCO and COCO-C as the clean and robust performance respectively. rPC~\cite{michaelis2019benchmarking} is used to to measure relative performance degradation under corruption.

\paragraph{Results} We compare the detectors with DAT, Det-AdvProp~\cite{chen2021robust} and vanilla training in Table~\ref{tab:main_result_det}. Although Det-AdvProp has been shown effective on EfficientDet with size larger than D0, we find on smaller detectors, the promotion of Det-AdvProp is subtle. On EfficientDet-Lite0 and YOLOv3-320, it even gets worse on COCO mAP than vanilla training. While our DAT achieves better result on both clean and corrupted input. EfficientDet-Lite0 with DAT get 27.3 and 17.89 mAP on COCO and COCO-C, resulting in 65.53\% Relative rPC. By comparison, vanilla YOLOv3 models have lower Relative rPC, showing it is more vulnerable than EfficientDet at same clean mAP. After equipped with DAT, YOLOv3-320 can get 20.55 mAP on COCO-C, leading to 7.75\% improvement on rPC.

\subsection{Ablations}

%%%%%%%%%%%%%%%%%%%%%%%%%%%%%%%%%%%%%%%%%%%%%
\begin{figure}
  \centering
  \includegraphics[width=1.\linewidth]{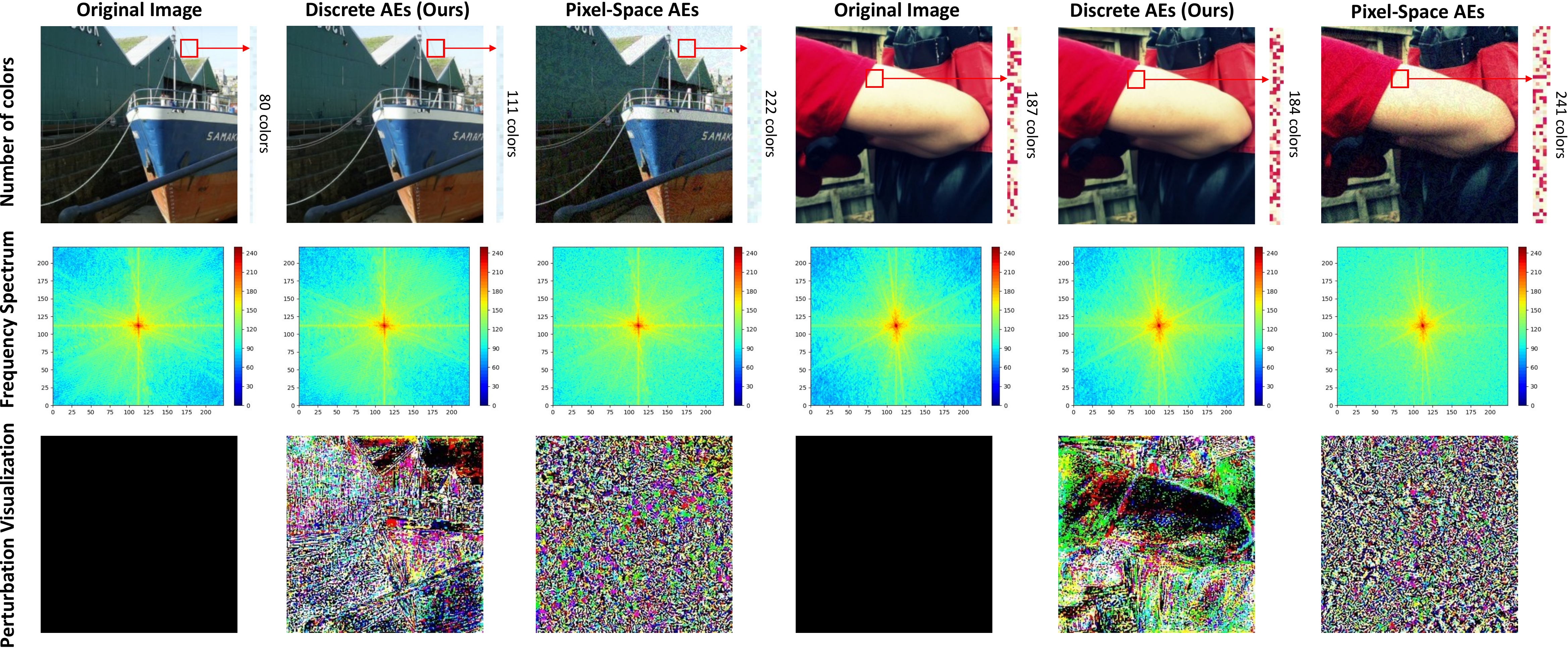}
  \vspace{-2em}
  \caption{Comparison of discrete perturbations and pixel-space perturbations.}
  \label{fig:adv_compare}
  \vspace{-1.5em}
\end{figure}
%%%%%%%%%%%%%%%%%%%%%%%%%%%%%%%%%%%%%%%%%%%%%

\paragraph{Discrete Perturbations vs. Pixel-Space Perturbations} We compare the proposed Discrete Adversarial Examples (DAEs) with traditional Pixel-space Adversarial Examples (PAEs) in Figure~\ref{fig:adv_compare}. DAEs have following three superior properties:
1) DAEs are more realistic. By calculating the number of colors~\cite{duan2021advdrop}, we find PAEs add more invalid colors, resulting in a noisy image. While DAEs have minor changes on the color numbers. The FID score of our DAEs is 14.65, which is lower than 65.18 of AEs; 2) DAEs have less high frequency component compared with AEs in frequency analysis; 3) Discrete perturbations are more structural, showing it attends to more important object locations. 

\paragraph{Impact of the Codebook in DAT} A general understanding is that the codebook with larger $K$ should have stronger representation power. As suggested in Table~\ref{tab:codebooktype}, when the $K$ is reduced from 16384 to 256, the FID of generated images increases 0.35. We show DAT on larger codebook size can achieve better generalization. However, the clean accuracy seems no improvement with increasing of $K$. We think the reason is that the reconstruction quality has already met the demand of DAT. And the improvement on clean accuracy is saturated when $K$ further increases.

\paragraph{Different Types of Discretizer $\mathcal{Q}$} Our work is based on the hypothesis of an ideal $\mathcal{Q}$ with no loss in discretization process. However, such perfect $\mathcal{Q}$ is not existing in practice. So we study if and how the ability of $\mathcal{Q}$ affects the performance of our DAT in Table~\ref{tab:codebooktype}. The results show DAT performs better by using discretizer with higher FID. We find clean accuracy declines 1.3\% on DALL-E, showing it is important for choosing a powerful discretizer. Types of pre-training datasets have little impact. We find the discretizer pre-trained on OpenImages~\cite{kuznetsova2020open} and transferred to discretize ImageNet images, can also achieve comparable results. 

\begin{table}\centering
\scriptsize
\begin{tabular}{l|ccccc|cccccc}
\toprule
 Models & Datasets & $f$ & $K$ & $d$ & FID & ImageNet & FGSM & A & C$\downarrow$ & R & Stylized \\
\midrule
 VQGAN~\cite{rombach2021high} & OpenImages~\cite{kuznetsova2020open} & 8 & 16384 & 4 & 1.14 & 76.52 & 30.66 & 4.38 & 74.16 & 41.90 & 10.80 \\
 VQGAN~\cite{rombach2021high} & OpenImages~\cite{kuznetsova2020open} & 16 & 16384 & 8 & 5.15 & 75.65 & 30.48 & 3.17 & 74.71 & 40.64 & 10.76 \\
 VQGAN~\cite{rombach2021high} & OpenImages~\cite{kuznetsova2020open} & 8 & 256 & 4 & 1.49 & 76.52 & 30.46 & 3.11 & 74.06 & 40.97 & 10.01  \\
 VQGAN~\cite{esser2021taming} & ImageNet~\cite{deng2009imagenet} & 16 & 16384 & 256 & 4.98 & 75.88 & 30.28 & 2.65 & 75.38 & 40.66 & 9.58 \\
 DALL-E~\cite{ramesh2021zero} & Private & 8 & 8192 & 128 & 32.01 & 75.23 & 30.17 & 3.97 & 74.26 & 40.13 & 10.65 \\
\bottomrule
\end{tabular}
\smallskip
\caption{Results of DAT based on different pretrained discretizer $\mathcal{Q}$. $f$ presents the downsampling factors. We use ResNet50 as base model and the subset of benchmarks in Sec~\ref{sec:4.1} for evaluation.}
\vspace{-3em}
\label{tab:codebooktype}
\end{table}

\begin{figure}[h!]
\vspace{-0.5em}
\centering
\vspace{-0.3em}
\subfigure[Clean Accuracy]{
\includegraphics[width=0.32\textwidth]{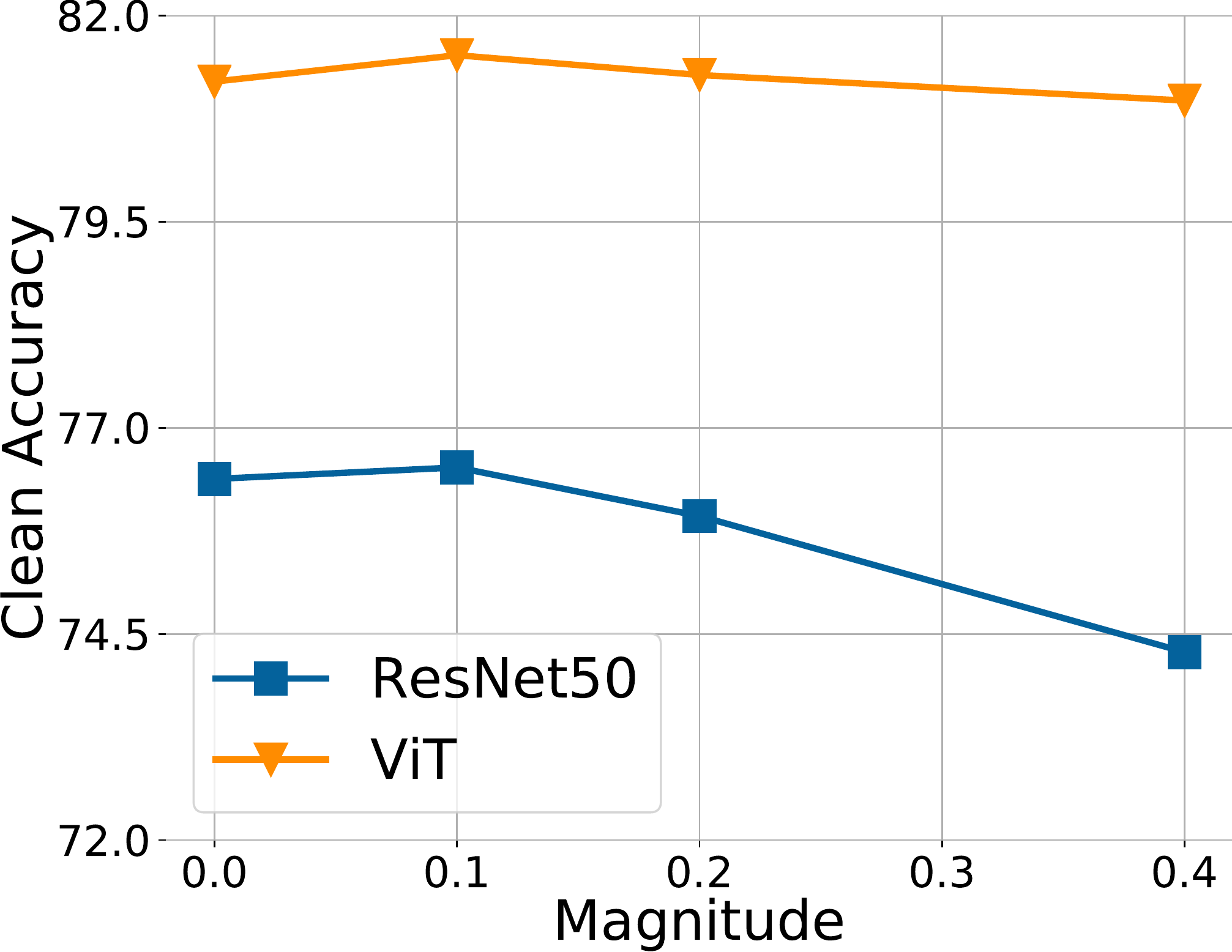}}
    \subfigure[Adversarial Robustness]{
\includegraphics[width=0.32\textwidth]{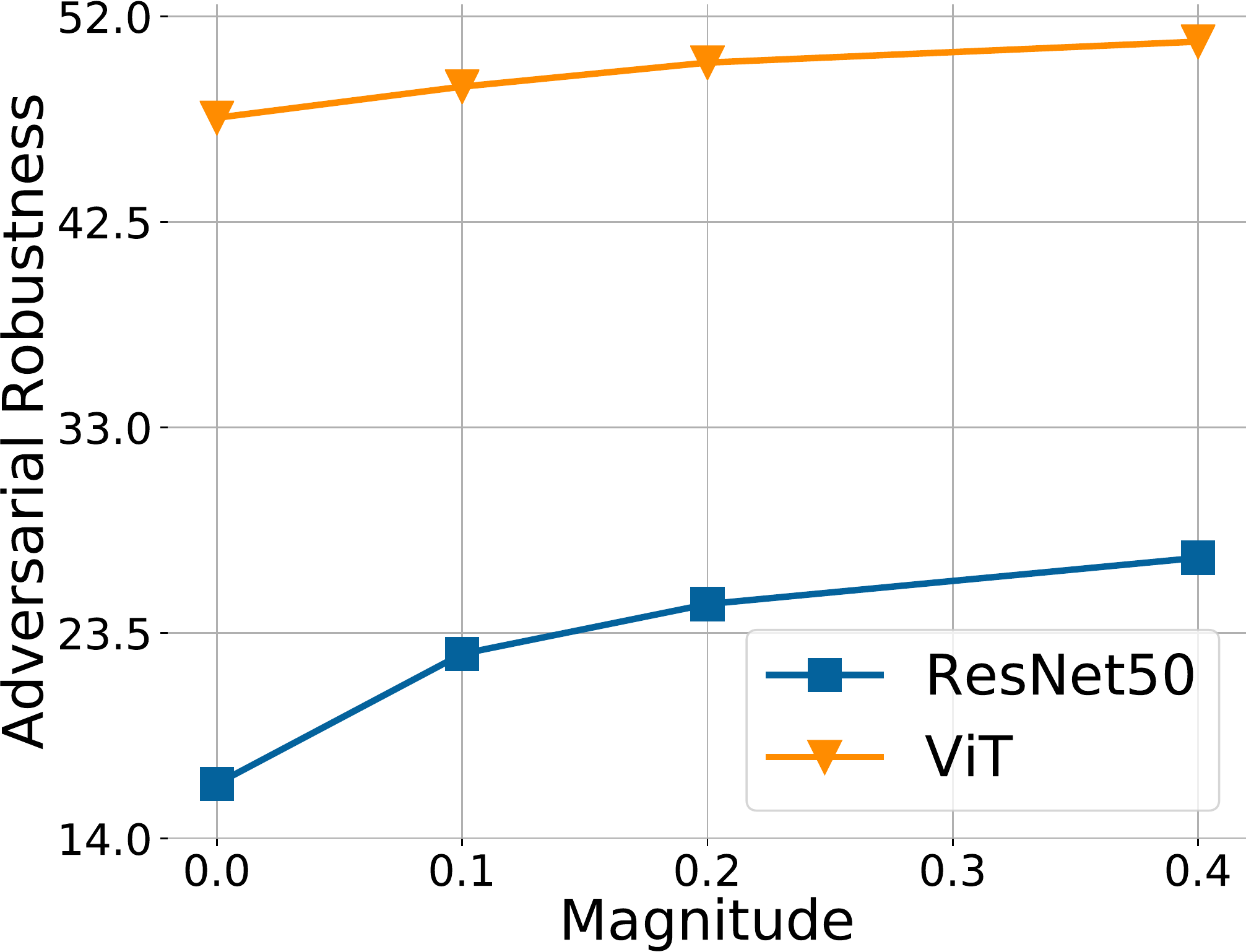}}
    \subfigure[OOD Robustness]{
\includegraphics[width=0.32\textwidth]{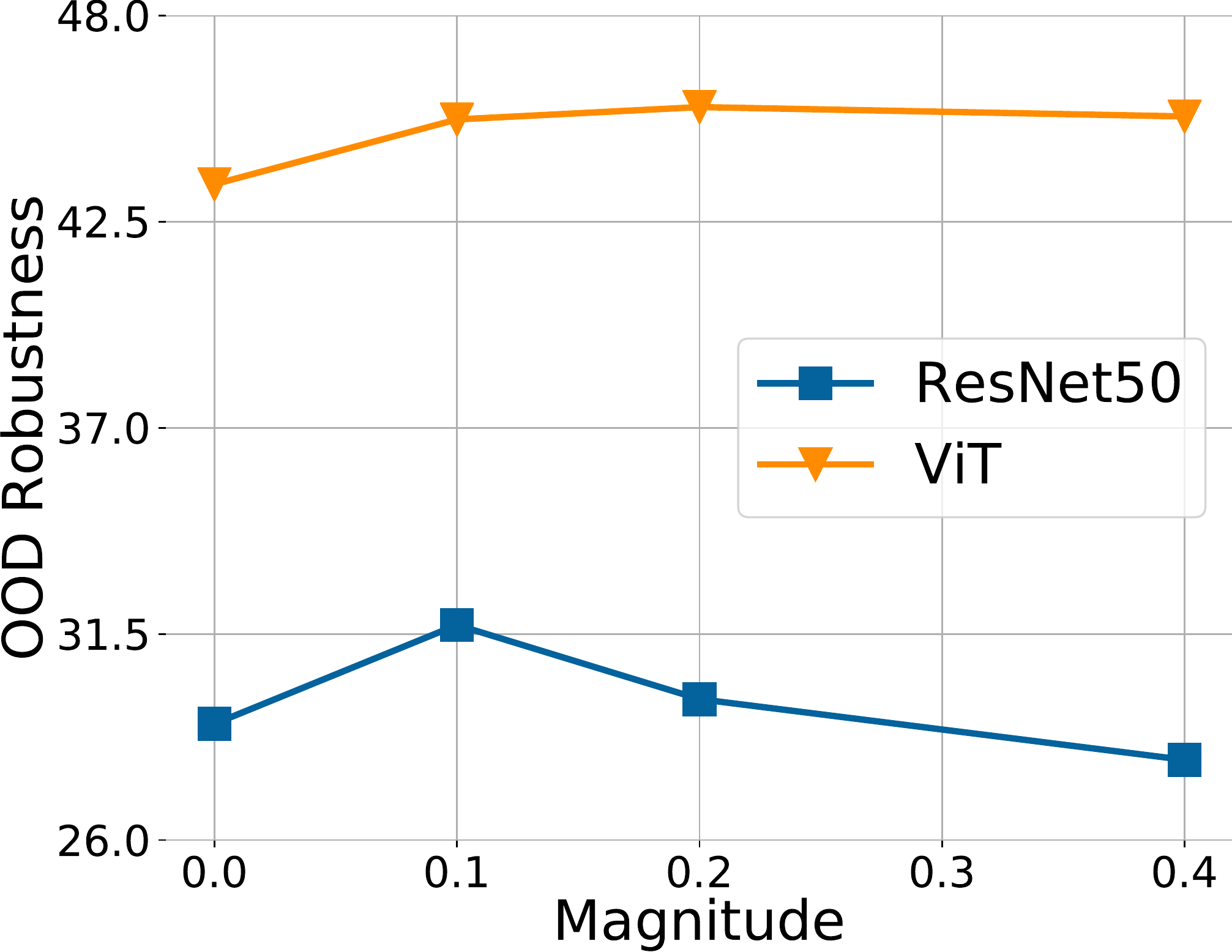}}
\vspace{-1em}
    \caption{The performance of DAT with different magnitude $\alpha$ in Eq~\ref{eq:gradient}.}
    \label{fig:ablation1}
    \vspace{-1.5em}
\end{figure}

\paragraph{The Performance on Different Magnitude $\alpha$}
We present the DAT results on ResNet50 and ViT with different magnitude $\alpha$ in Figure~\ref{fig:ablation1}. $\alpha=0$ means the model is only trained on images augmented by VQGAN with no adversarial training process. We show DAT with $\alpha=0$ makes the models have high clean accuracy but the lowest robustness. With the increase of $\alpha$, there is still a robustness and accuracy trade-off. DAT with $\alpha=0.4$ on ResNet50 achieves the best adversarial robustness but generalization is getting worse. By contrast, we surprisingly find ViT has the lower sensibility on $\alpha$. Even if ViT is trained on DAT with $\alpha=0.4$, there is still no great drop of clean and OOD accuracy. We suspect that ViT is more suitable for training on discrete examples, and the strong modeling ability makes ViT greater than ResNet50. 

%%%%%%%%%%%%%%%%%%%%%%%%%%%%%%%%%%%%%%%%%%%
%%%%%%%%%%%%%%%%%%%%%%%%%%%%%%%%%%%%%%%%%%%

\begin{figure}[h!]
\vspace{-0.7em}
\centering
\subfigure[AutoAttack~\cite{croce2020reliable}]{
\includegraphics[width=0.32\textwidth]{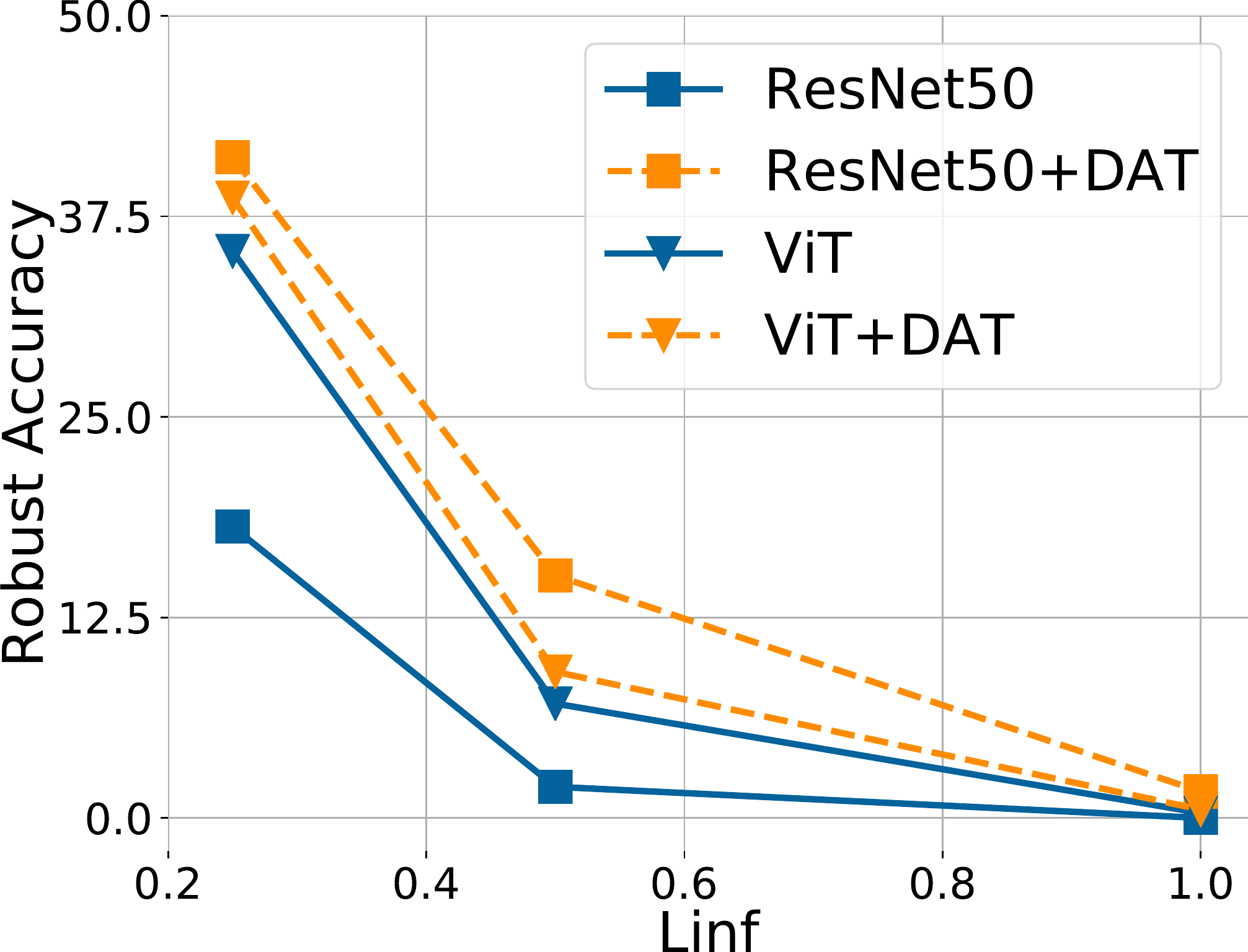}}
    \subfigure[AdvDrop~\cite{duan2021advdrop}]{
\includegraphics[width=0.32\textwidth]{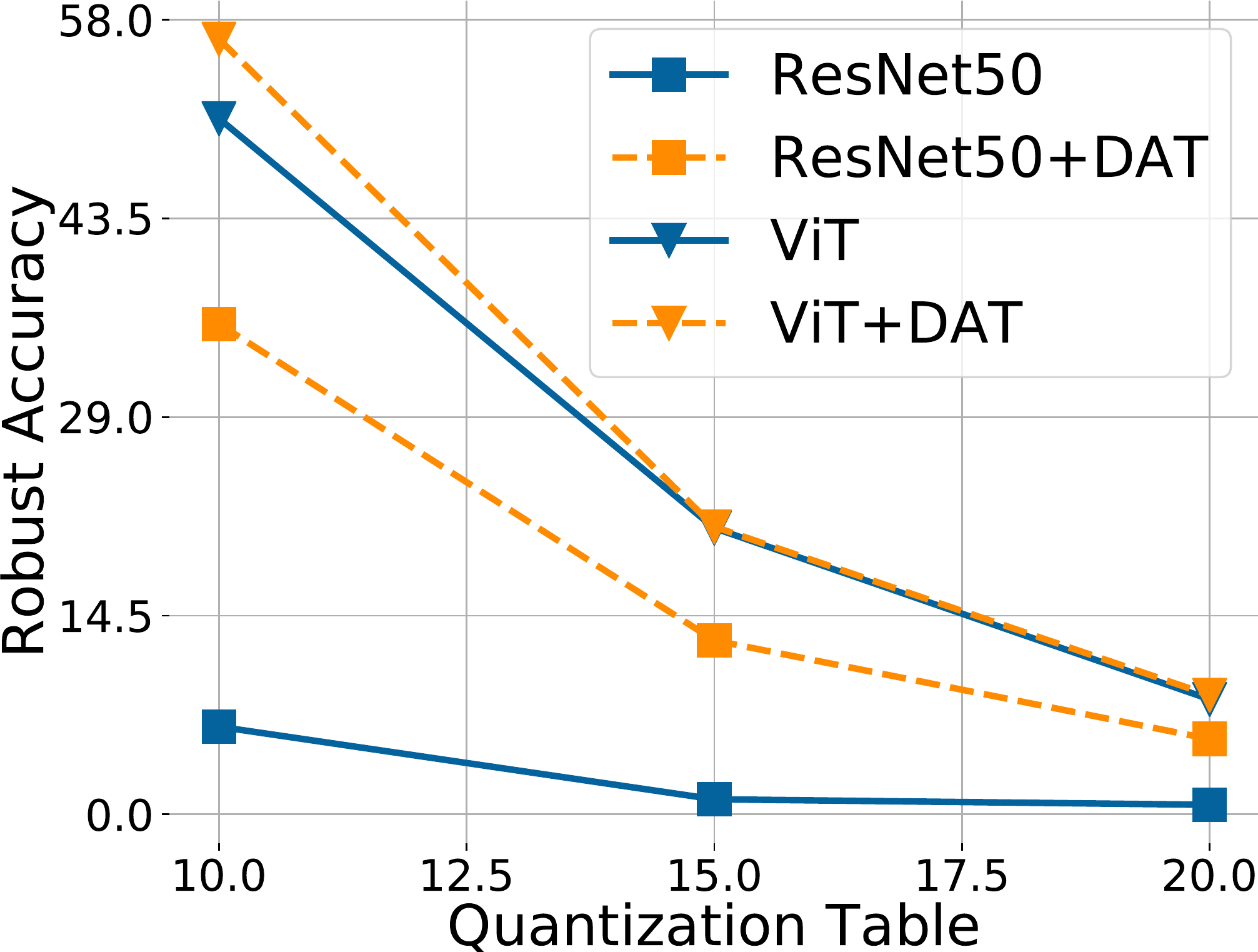}}
    \subfigure[PerC-Adversarial~\cite{zhao2020towards}]{
\includegraphics[width=0.32\textwidth]{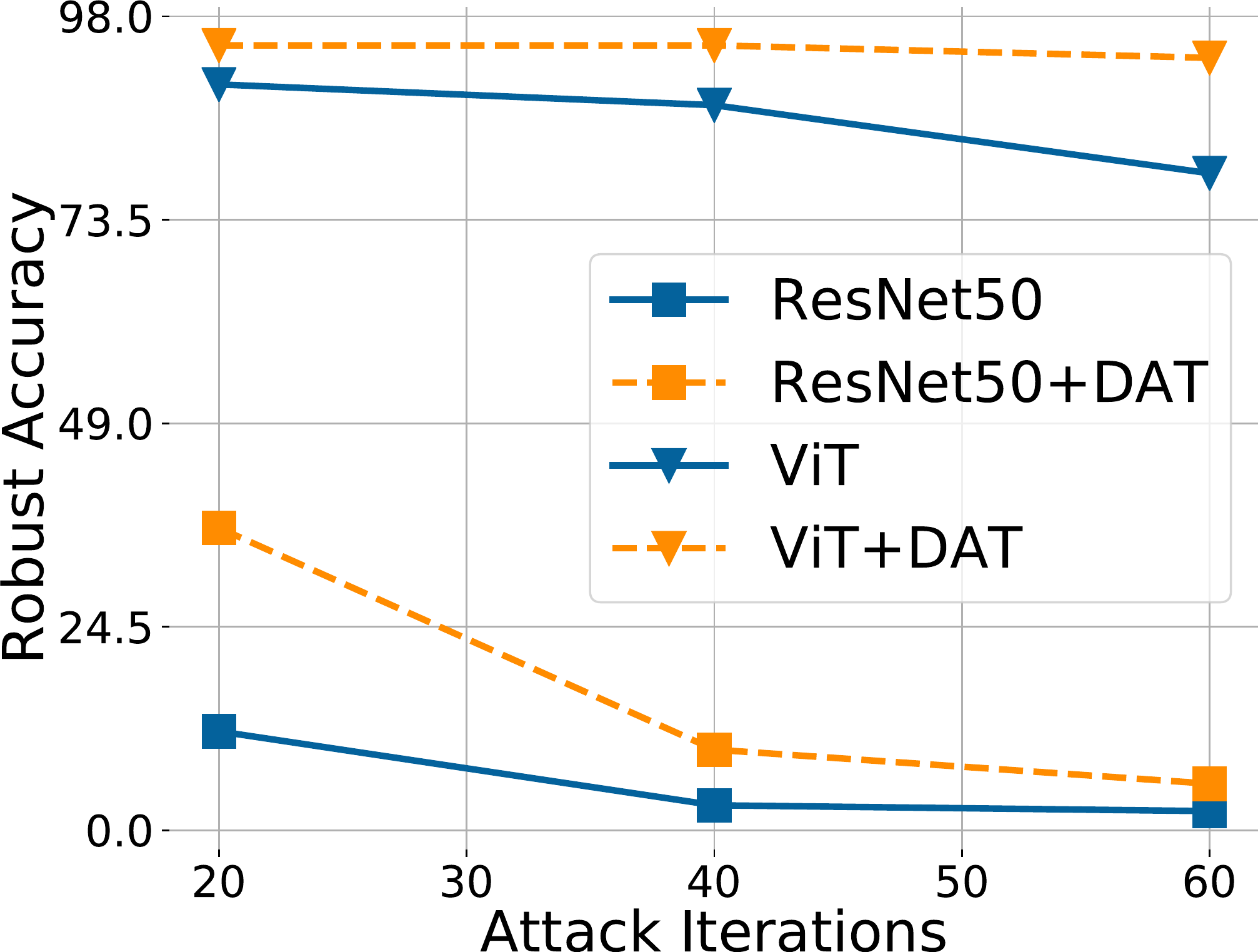}}
\vspace{-1em}
    \caption{The adversarial robustness test under other stronger attackers.}
    \label{fig:ablation2}
    \vspace{-1.5em}
\end{figure}

\paragraph{Results on Stronger Attacks} In the main experiment, only one-step FGSM is used for examining the robustness under white-box adversarial attacks. To give more comprehensive evaluation, we additionally test our DAT under stronger AutoAttack~\cite{croce2020reliable}, and two unrestricted attacks named AdvDrop~\cite{duan2021advdrop} and PerC-Adversarial~\cite{zhao2020towards}. The result is shown in Figure~\ref{fig:ablation2}. DAT brings the improvement of robustness under all three attackers. ViT trained with DAT achieved extremely high robust accuracy on PerC-Adversarial, which suggests DAT also effects well in defense against unrestricted adversarial attacks. More details can be referred to Appendix C.

%%%%%%%%%%%%%%%%%%%%%%%%%%%%%%%%%%%%%%%%%%%
%%%%%%%%%%%%%%%%%%%%%%%%%%%%%%%%%%%%%%%%%%%

\section{Limitations and Conclusions}
In this paper, we find transferring NLP-style adversarial training to vision models can enhance the learned visual representation effectively. We propose Discrete Adversarial Training (DAT), where images are presented as discrete visual words by VQGAN, and the model is training on examples which have the adversarially altered discrete visual representation. DAT needs not to modify the model architecture and works for both CNNs and ViTs across multiple tasks. DAT reports the state-of-the-art robustness on ImageNet-C and Stylized-ImageNet, exhibiting strong generalization. However, DAT still costs increased training time, this limitation also holds for any adversarial training. The strict assumption of an ideal discretizer is another potential limitation, despite DAT has used powerful VQGAN model to approach this assumption. The effect of DAT is empirically studied without deeper theoretical explanation. All the above limitations are remained as the future optimization direction.

{
\small
\bibliographystyle{unsrt}
\bibliography{main}

\begin{thebibliography}{10}

\bibitem{Szegedy2014IntriguingPO}
Christian Szegedy, Wojciech Zaremba, Ilya Sutskever, Joan Bruna, D.~Erhan,
  Ian~J. Goodfellow, and Rob Fergus.
\newblock Intriguing properties of neural networks.
\newblock In {\em International Conference on Learning Representations}, 2019.

\bibitem{Hendrycks2019BenchmarkingNN}
Dan Hendrycks and Thomas~G. Dietterich.
\newblock Benchmarking neural network robustness to common corruptions and
  perturbations.
\newblock In {\em International Conference on Learning Representations}, 2019.

\bibitem{geirhos2018imagenet}
Robert Geirhos, Patricia Rubisch, Claudio Michaelis, Matthias Bethge, Felix~A
  Wichmann, and Wieland Brendel.
\newblock Imagenet-trained cnns are biased towards texture; increasing shape
  bias improves accuracy and robustness.
\newblock In {\em International Conference on Learning Representations}, 2018.

\bibitem{shamsabadi2020colorfool}
Ali~Shahin Shamsabadi, Ricardo Sanchez-Matilla, and Andrea Cavallaro.
\newblock Colorfool: Semantic adversarial colorization.
\newblock In {\em Proceedings of the IEEE/CVF Conference on Computer Vision and
  Pattern Recognition}, pages 1151--1160, 2020.

\bibitem{madry2018towards}
Aleksander Madry, Aleksandar Makelov, Ludwig Schmidt, Dimitris Tsipras, and
  Adrian Vladu.
\newblock Towards deep learning models resistant to adversarial attacks.
\newblock In {\em International Conference on Learning Representations}, 2018.

\bibitem{engstrom2019learning}
Logan Engstrom, Andrew Ilyas, Shibani Santurkar, Dimitris Tsipras, Brandon
  Tran, and Aleksander Madry.
\newblock Learning perceptually-aligned representations via adversarial
  robustness.
\newblock {\em arXiv preprint arXiv:1906.00945}, 2(3):5, 2019.

\bibitem{ross2018improving}
Andrew Ross and Finale Doshi-Velez.
\newblock Improving the adversarial robustness and interpretability of deep
  neural networks by regularizing their input gradients.
\newblock In {\em Proceedings of the AAAI Conference on Artificial
  Intelligence}, volume~32, 2018.

\bibitem{ilyas2019adversarial}
Andrew Ilyas, Shibani Santurkar, Dimitris Tsipras, Logan Engstrom, Brandon
  Tran, and Aleksander Madry.
\newblock Adversarial examples are not bugs, they are features.
\newblock {\em Advances in neural information processing systems}, 32, 2019.

\bibitem{salman2020adversarially}
Hadi Salman, Andrew Ilyas, Logan Engstrom, Ashish Kapoor, and Aleksander Madry.
\newblock Do adversarially robust imagenet models transfer better?
\newblock {\em Advances in Neural Information Processing Systems},
  33:3533--3545, 2020.

\bibitem{wen2020towards}
Yuxin Wen, Shuai Li, and Kui Jia.
\newblock Towards understanding the regularization of adversarial robustness on
  neural networks.
\newblock In {\em International Conference on Machine Learning}, pages
  10225--10235. PMLR, 2020.

\bibitem{zhu2019freelb}
Chen Zhu, Yu~Cheng, Zhe Gan, Siqi Sun, Tom Goldstein, and Jingjing Liu.
\newblock Freelb: Enhanced adversarial training for natural language
  understanding.
\newblock In {\em International Conference on Learning Representations}, 2019.

\bibitem{ivgi2021achieving}
Maor Ivgi and Jonathan Berant.
\newblock Achieving model robustness through discrete adversarial training.
\newblock In {\em Proceedings of the 2021 Conference on Empirical Methods in
  Natural Language Processing}, pages 1529--1544, 2021.

\bibitem{esser2021taming}
Patrick Esser, Robin Rombach, and Bjorn Ommer.
\newblock Taming transformers for high-resolution image synthesis.
\newblock In {\em Proceedings of the IEEE/CVF Conference on Computer Vision and
  Pattern Recognition}, pages 12873--12883, 2021.

\bibitem{jin2020bert}
Di~Jin, Zhijing Jin, Joey~Tianyi Zhou, and Peter Szolovits.
\newblock Is bert really robust? a strong baseline for natural language attack
  on text classification and entailment.
\newblock In {\em Proceedings of the AAAI conference on artificial
  intelligence}, volume~34, pages 8018--8025, 2020.

\bibitem{zang2020word}
Yuan Zang, Fanchao Qi, Chenghao Yang, Zhiyuan Liu, Meng Zhang, Qun Liu, and
  Maosong Sun.
\newblock Word-level textual adversarial attacking as combinatorial
  optimization.
\newblock In {\em Proceedings of the 58th Annual Meeting of the Association for
  Computational Linguistics}, pages 6066--6080, 2020.

\bibitem{alzantot2018generating}
Moustafa Alzantot, Yash Sharma, Ahmed Elgohary, Bo-Jhang Ho, Mani Srivastava,
  and Kai-Wei Chang.
\newblock Generating natural language adversarial examples.
\newblock In {\em Proceedings of the 2018 Conference on Empirical Methods in
  Natural Language Processing}, pages 2890--2896, 2018.

\bibitem{duan2021advdrop}
Ranjie Duan, Yuefeng Chen, Dantong Niu, Yun Yang, A~Kai Qin, and Yuan He.
\newblock Advdrop: Adversarial attack to dnns by dropping information.
\newblock In {\em Proceedings of the IEEE/CVF International Conference on
  Computer Vision}, pages 7506--7515, 2021.

\bibitem{zhao2020towards}
Zhengyu Zhao, Zhuoran Liu, and Martha Larson.
\newblock Towards large yet imperceptible adversarial image perturbations with
  perceptual color distance.
\newblock In {\em Proceedings of the IEEE/CVF Conference on Computer Vision and
  Pattern Recognition}, pages 1039--1048, 2020.

\bibitem{he2021masked}
Kaiming He, Xinlei Chen, Saining Xie, Yanghao Li, Piotr Doll{\'a}r, and Ross
  Girshick.
\newblock Masked autoencoders are scalable vision learners.
\newblock In {\em Proceedings of the IEEE/CVF International Conference on
  Computer Vision}, 2022.

\bibitem{dosovitskiy2020image}
Alexey Dosovitskiy, Lucas Beyer, Alexander Kolesnikov, Dirk Weissenborn,
  Xiaohua Zhai, Thomas Unterthiner, Mostafa Dehghani, Matthias Minderer, Georg
  Heigold, Sylvain Gelly, et~al.
\newblock An image is worth 16x16 words: Transformers for image recognition at
  scale.
\newblock In {\em International Conference on Learning Representations}, 2020.

\bibitem{tsipras2018robustness}
Dimitris Tsipras, Shibani Santurkar, Logan Engstrom, Alexander Turner, and
  Aleksander Madry.
\newblock Robustness may be at odds with accuracy.
\newblock In {\em International Conference on Learning Representations}, 2018.

\bibitem{zhang2019theoretically}
Hongyang Zhang, Yaodong Yu, Jiantao Jiao, Eric Xing, Laurent El~Ghaoui, and
  Michael Jordan.
\newblock Theoretically principled trade-off between robustness and accuracy.
\newblock In {\em International conference on machine learning}, pages
  7472--7482. PMLR, 2019.

\bibitem{raghunathan2020understanding}
Aditi Raghunathan, Sang~Michael Xie, Fanny Yang, John Duchi, and Percy Liang.
\newblock Understanding and mitigating the tradeoff between robustness and
  accuracy.
\newblock In {\em International Conference on Machine Learning}, pages
  7909--7919. PMLR, 2020.

\bibitem{rade2021reducing}
Rahul Rade and Seyed-Mohsen Moosavi-Dezfooli.
\newblock Reducing excessive margin to achieve a better accuracy vs. robustness
  trade-off.
\newblock In {\em International Conference on Learning Representations}, 2021.

\bibitem{lamb2019interpolated}
Alex Lamb, Vikas Verma, Juho Kannala, and Yoshua Bengio.
\newblock Interpolated adversarial training: Achieving robust neural networks
  without sacrificing too much accuracy.
\newblock In {\em Proceedings of the 12th ACM Workshop on Artificial
  Intelligence and Security}, pages 95--103, 2019.

\bibitem{xie2020adversarial}
Cihang Xie, Mingxing Tan, Boqing Gong, Jiang Wang, Alan~L Yuille, and Quoc~V
  Le.
\newblock Adversarial examples improve image recognition.
\newblock In {\em Proceedings of the IEEE/CVF Conference on Computer Vision and
  Pattern Recognition}, pages 819--828, 2020.

\bibitem{mei2021fast}
Jieru Mei, Yucheng Han, Yutong Bai, Yixiao Zhang, Yingwei Li, Xianhang Li, Alan
  Yuille, and Cihang Xie.
\newblock Fast advprop.
\newblock In {\em International Conference on Learning Representations}, 2021.

\bibitem{herrmann2021pyramid}
Charles Herrmann, Kyle Sargent, Lu~Jiang, Ramin Zabih, Huiwen Chang, Ce~Liu,
  Dilip Krishnan, and Deqing Sun.
\newblock Pyramid adversarial training improves vit performance.
\newblock {\em arXiv preprint arXiv:2111.15121}, 2021.

\bibitem{gan2020large}
Zhe Gan, Yen-Chun Chen, Linjie Li, Chen Zhu, Yu~Cheng, and Jingjing Liu.
\newblock Large-scale adversarial training for vision-and-language
  representation learning.
\newblock In {\em NeurIPS}, 2020.

\bibitem{gokhale2021attribute}
Tejas Gokhale, Rushil Anirudh, Bhavya Kailkhura, Jayaraman~J Thiagarajan,
  Chitta Baral, and Yezhou Yang.
\newblock Attribute-guided adversarial training for robustness to natural
  perturbations.
\newblock In {\em Proceedings of the AAAI Conference on Artificial
  Intelligence}, volume~35, pages 7574--7582, 2021.

\bibitem{zhang2019adversarial}
Xinyu Zhang, Qiang Wang, Jian Zhang, and Zhao Zhong.
\newblock Adversarial autoaugment.
\newblock In {\em International Conference on Learning Representations}, 2019.

\bibitem{wang2021augmax}
Haotao Wang, Chaowei Xiao, Jean Kossaifi, Zhiding Yu, Anima Anandkumar, and
  Zhangyang Wang.
\newblock Augmax: Adversarial composition of random augmentations for robust
  training.
\newblock {\em Advances in Neural Information Processing Systems}, 34, 2021.

\bibitem{calian2021defending}
Dan~Andrei Calian, Florian Stimberg, Olivia Wiles, Sylvestre-Alvise Rebuffi,
  Andr{\'a}s Gy{\"o}rgy, Timothy~A Mann, and Sven Gowal.
\newblock Defending against image corruptions through adversarial
  augmentations.
\newblock In {\em International Conference on Learning Representations}, 2021.

\bibitem{hendrycks2019augmix}
Dan Hendrycks, Norman Mu, Ekin~Dogus Cubuk, Barret Zoph, Justin Gilmer, and
  Balaji Lakshminarayanan.
\newblock Augmix: A simple data processing method to improve robustness and
  uncertainty.
\newblock In {\em International Conference on Learning Representations}, 2019.

\bibitem{gong2021maxup}
Chengyue Gong, Tongzheng Ren, Mao Ye, and Qiang Liu.
\newblock Maxup: Lightweight adversarial training with data augmentation
  improves neural network training.
\newblock In {\em Proceedings of the IEEE/CVF Conference on Computer Vision and
  Pattern Recognition}, pages 2474--2483, 2021.

\bibitem{sivic2003video}
Josef Sivic and Andrew Zisserman.
\newblock Video google: A text retrieval approach to object matching in videos.
\newblock In {\em Computer Vision, IEEE International Conference on}, volume~3,
  pages 1470--1470. IEEE Computer Society, 2003.

\bibitem{csurka2004visual}
Gabriella Csurka, Christopher Dance, Lixin Fan, Jutta Willamowski, and
  C{\'e}dric Bray.
\newblock Visual categorization with bags of keypoints.
\newblock In {\em Workshop on statistical learning in computer vision, ECCV},
  volume~1, pages 1--2. Prague, 2004.

\bibitem{van2017neural}
Aaron Van Den~Oord, Oriol Vinyals, et~al.
\newblock Neural discrete representation learning.
\newblock {\em Advances in neural information processing systems}, 30, 2017.

\bibitem{bao2021beit}
Hangbo Bao, Li~Dong, and Furu Wei.
\newblock Beit: Bert pre-training of image transformers.
\newblock In {\em International Conference on Learning Representations}, 2022.

\bibitem{zhou2021ibot}
Jinghao Zhou, Chen Wei, Huiyu Wang, Wei Shen, Cihang Xie, Alan Yuille, and Tao
  Kong.
\newblock ibot: Image bert pre-training with online tokenizer.
\newblock {\em arXiv preprint arXiv:2111.07832}, 2021.

\bibitem{mao2021discrete}
Chengzhi Mao, Lu~Jiang, Mostafa Dehghani, Carl Vondrick, Rahul Sukthankar, and
  Irfan Essa.
\newblock Discrete representations strengthen vision transformer robustness.
\newblock In {\em International Conference on Learning Representations}, 2022.

\bibitem{rombach2021high}
Robin Rombach, Andreas Blattmann, Dominik Lorenz, Patrick Esser, and Bj{\"o}rn
  Ommer.
\newblock High-resolution image synthesis with latent diffusion models.
\newblock {\em arXiv preprint arXiv:2112.10752}, 2021.

\bibitem{bengio2013estimating}
Yoshua Bengio, Nicholas L{\'e}onard, and Aaron Courville.
\newblock Estimating or propagating gradients through stochastic neurons for
  conditional computation.
\newblock {\em arXiv preprint arXiv:1308.3432}, 2013.

\bibitem{yin2019understanding}
Penghang Yin, Jiancheng Lyu, Shuai Zhang, Stanley Osher, Yingyong Qi, and Jack
  Xin.
\newblock Understanding straight-through estimator in training activation
  quantized neural nets.
\newblock {\em arXiv preprint arXiv:1903.05662}, 2019.

\bibitem{ioffe2015batch}
Sergey Ioffe and Christian Szegedy.
\newblock Batch normalization: Accelerating deep network training by reducing
  internal covariate shift.
\newblock In {\em International conference on machine learning}, pages
  448--456. PMLR, 2015.

\bibitem{he2016deep}
Kaiming He, Xiangyu Zhang, Shaoqing Ren, and Jian Sun.
\newblock Deep residual learning for image recognition.
\newblock In {\em Proceedings of the IEEE conference on computer vision and
  pattern recognition}, pages 770--778, 2016.

\bibitem{hendrycks2021many}
Dan Hendrycks, Steven Basart, Norman Mu, Saurav Kadavath, Frank Wang, Evan
  Dorundo, Rahul Desai, Tyler Zhu, Samyak Parajuli, Mike Guo, et~al.
\newblock The many faces of robustness: A critical analysis of
  out-of-distribution generalization.
\newblock In {\em Proceedings of the IEEE/CVF International Conference on
  Computer Vision}, pages 8340--8349, 2021.

\bibitem{steiner2021train}
Andreas Steiner, Alexander Kolesnikov, Xiaohua Zhai, Ross Wightman, Jakob
  Uszkoreit, and Lucas Beyer.
\newblock How to train your vit? data, augmentation, and regularization in
  vision transformers.
\newblock {\em arXiv preprint arXiv:2106.10270}, 2021.

\bibitem{goodfellow2014explaining}
Ian~J Goodfellow, Jonathon Shlens, and Christian Szegedy.
\newblock Explaining and harnessing adversarial examples.
\newblock {\em arXiv preprint arXiv:1412.6572}, 2014.

\bibitem{chen2020universal}
Sizhe Chen, Zhengbao He, Chengjin Sun, Jie Yang, and Xiaolin Huang.
\newblock Universal adversarial attack on attention and the resulting dataset
  damagenet.
\newblock {\em IEEE Transactions on Pattern Analysis and Machine Intelligence},
  2020.

\bibitem{hendrycks2021natural}
Dan Hendrycks, Kevin Zhao, Steven Basart, Jacob Steinhardt, and Dawn Song.
\newblock Natural adversarial examples.
\newblock In {\em Proceedings of the IEEE/CVF Conference on Computer Vision and
  Pattern Recognition}, pages 15262--15271, 2021.

\bibitem{hendrycks2018benchmarking}
Dan Hendrycks and Thomas Dietterich.
\newblock Benchmarking neural network robustness to common corruptions and
  perturbations.
\newblock In {\em International Conference on Learning Representations}, 2018.

\bibitem{wang2019learning}
Haohan Wang, Songwei Ge, Zachary Lipton, and Eric~P Xing.
\newblock Learning robust global representations by penalizing local predictive
  power.
\newblock {\em Advances in Neural Information Processing Systems}, 32, 2019.

\bibitem{zhang2018mixup}
Hongyi Zhang, Moustapha Cisse, Yann~N Dauphin, and David Lopez-Paz.
\newblock mixup: Beyond empirical risk minimization.
\newblock In {\em International Conference on Learning Representations}, 2018.

\bibitem{yun2019cutmix}
Sangdoo Yun, Dongyoon Han, Seong~Joon Oh, Sanghyuk Chun, Junsuk Choe, and
  Youngjoon Yoo.
\newblock Cutmix: Regularization strategy to train strong classifiers with
  localizable features.
\newblock In {\em Proceedings of the IEEE/CVF international conference on
  computer vision}, pages 6023--6032, 2019.

\bibitem{cubuk2020randaugment}
Ekin~D Cubuk, Barret Zoph, Jonathon Shlens, and Quoc~V Le.
\newblock Randaugment: Practical automated data augmentation with a reduced
  search space.
\newblock In {\em Proceedings of the IEEE/CVF Conference on Computer Vision and
  Pattern Recognition Workshops}, pages 702--703, 2020.

\bibitem{li2020shape}
Yingwei Li, Qihang Yu, Mingxing Tan, Jieru Mei, Peng Tang, Wei Shen, Alan
  Yuille, et~al.
\newblock Shape-texture debiased neural network training.
\newblock In {\em International Conference on Learning Representations}, 2020.

\bibitem{chen2020simple}
Ting Chen, Simon Kornblith, Mohammad Norouzi, and Geoffrey Hinton.
\newblock A simple framework for contrastive learning of visual
  representations.
\newblock In {\em International conference on machine learning}, pages
  1597--1607. PMLR, 2020.

\bibitem{chen2021exploring}
Xinlei Chen and Kaiming He.
\newblock Exploring simple siamese representation learning.
\newblock In {\em Proceedings of the IEEE/CVF Conference on Computer Vision and
  Pattern Recognition}, pages 15750--15758, 2021.

\bibitem{chen2021empirical}
Xinlei Chen, Saining Xie, and Kaiming He.
\newblock An empirical study of training self-supervised vision transformers.
\newblock In {\em Proceedings of the IEEE/CVF International Conference on
  Computer Vision}, pages 9640--9649, 2021.

\bibitem{ericsson2021well}
Linus Ericsson, Henry Gouk, and Timothy~M Hospedales.
\newblock How well do self-supervised models transfer?
\newblock In {\em Proceedings of the IEEE/CVF Conference on Computer Vision and
  Pattern Recognition}, pages 5414--5423, 2021.

\bibitem{kim2020adversarial}
Minseon Kim, Jihoon Tack, and Sung~Ju Hwang.
\newblock Adversarial self-supervised contrastive learning.
\newblock {\em Advances in Neural Information Processing Systems},
  33:2983--2994, 2020.

\bibitem{everingham2010pascal}
Mark Everingham, Luc Van~Gool, Christopher~KI Williams, John Winn, and Andrew
  Zisserman.
\newblock The pascal visual object classes (voc) challenge.
\newblock {\em International journal of computer vision}, 88(2):303--338, 2010.

\bibitem{zhou2017scene}
Bolei Zhou, Hang Zhao, Xavier Puig, Sanja Fidler, Adela Barriuso, and Antonio
  Torralba.
\newblock Scene parsing through ade20k dataset.
\newblock In {\em Proceedings of the IEEE conference on computer vision and
  pattern recognition}, pages 633--641, 2017.

\bibitem{tan2020efficientdet}
Mingxing Tan, Ruoming Pang, and Quoc~V Le.
\newblock Efficientdet: Scalable and efficient object detection.
\newblock In {\em Proceedings of the IEEE/CVF conference on computer vision and
  pattern recognition}, pages 10781--10790, 2020.

\bibitem{redmon2018yolov3}
Joseph Redmon and Ali Farhadi.
\newblock Yolov3: An incremental improvement.
\newblock {\em arXiv preprint arXiv:1804.02767}, 2018.

\bibitem{chen2021robust}
Xiangning Chen, Cihang Xie, Mingxing Tan, Li~Zhang, Cho-Jui Hsieh, and Boqing
  Gong.
\newblock Robust and accurate object detection via adversarial learning.
\newblock In {\em Proceedings of the IEEE/CVF Conference on Computer Vision and
  Pattern Recognition}, pages 16622--16631, 2021.

\bibitem{lin2014microsoft}
Tsung-Yi Lin, Michael Maire, Serge Belongie, James Hays, Pietro Perona, Deva
  Ramanan, Piotr Doll{\'a}r, and C~Lawrence Zitnick.
\newblock Microsoft coco: Common objects in context.
\newblock In {\em European conference on computer vision}, pages 740--755.
  Springer, 2014.

\bibitem{michaelis2019benchmarking}
Claudio Michaelis, Benjamin Mitzkus, Robert Geirhos, Evgenia Rusak, Oliver
  Bringmann, Alexander~S Ecker, Matthias Bethge, and Wieland Brendel.
\newblock Benchmarking robustness in object detection: Autonomous driving when
  winter is coming.
\newblock {\em arXiv preprint arXiv:1907.07484}, 2019.

\bibitem{kuznetsova2020open}
Alina Kuznetsova, Hassan Rom, Neil Alldrin, Jasper Uijlings, Ivan Krasin, Jordi
  Pont-Tuset, Shahab Kamali, Stefan Popov, Matteo Malloci, Alexander
  Kolesnikov, et~al.
\newblock The open images dataset v4.
\newblock {\em International Journal of Computer Vision}, 128(7):1956--1981,
  2020.

\bibitem{deng2009imagenet}
Jia Deng, Wei Dong, Richard Socher, Li-Jia Li, Kai Li, and Li~Fei-Fei.
\newblock Imagenet: A large-scale hierarchical image database.
\newblock In {\em 2009 IEEE conference on computer vision and pattern
  recognition}, pages 248--255. Ieee, 2009.

\bibitem{ramesh2021zero}
Aditya Ramesh, Mikhail Pavlov, Gabriel Goh, Scott Gray, Chelsea Voss, Alec
  Radford, Mark Chen, and Ilya Sutskever.
\newblock Zero-shot text-to-image generation.
\newblock In {\em International Conference on Machine Learning}, pages
  8821--8831. PMLR, 2021.

\bibitem{croce2020reliable}
Francesco Croce and Matthias Hein.
\newblock Reliable evaluation of adversarial robustness with an ensemble of
  diverse parameter-free attacks.
\newblock In {\em International conference on machine learning}, pages
  2206--2216. PMLR, 2020.

\bibitem{goodfellow2014generative}
Ian Goodfellow, Jean Pouget-Abadie, Mehdi Mirza, Bing Xu, David Warde-Farley,
  Sherjil Ozair, Aaron Courville, and Yoshua Bengio.
\newblock Generative adversarial nets.
\newblock {\em Advances in neural information processing systems}, 27, 2014.

\end{thebibliography}
}

\clearpage

\appendix

The Appendix is organized as follows. Appendix~\ref{app: 1} extends the discussion on training of VQGAN. Appendix~\ref{app: 2} presents more experimental results including comparison of robust training strategies, effect analysis on different perturbations, images corruptions and inference with the discretizer $\mathcal{Q}$. Besides, we analyse the training budget of our DAT. Appendix~\ref{app: 3} discusses the implementation details of the three attackers in ablation experiments. Finally we do some visualizations in Appendix~\ref{app: 4}. 

\section{Details of Training VQGAN}
\label{app: 1}
We add more details about how VQGAN can be trained to discretize a continous image. The forward pass of VQGAN has been introduced in Sec~\ref{sec:3.2.1}. Suppose a reconstructed image $\hat{x}$, the training objective for VQGAN is defined between $\hat{x}$ and $x$ as:
\begin{align}
    &\mathcal{L}_{\text{VQGAN}}=\min_{\mathsf{Enc},\mathsf{Dec},\mathcal{Z}} \max_{D}\mathbb{E}_{x\sim p(x)}[
    \mathcal{L}_{\text{VQ}}(\mathsf{Enc},\mathsf{Dec},\mathcal{Z})+\mathcal{L}_{\text{GAN}}(\{\mathsf{Enc},\mathsf{Dec},\mathcal{Z}\}, D)]
    \\& \mathcal{L}_{\text{VQ}}(\mathsf{Enc},\mathsf{Dec},\mathcal{Z})= \|x-\hat{x} \|_{\text{precept}} + \|\text{sg}[\mathsf{Enc}(x)]-v_{\textbf{q}} \|_{2}^2+\|\text{sg}[v_{\textbf{q}}]-\mathsf{Enc}(x) \|_{2}^2
    \\& \mathcal{L}_{\text{GAN}}(\{\mathsf{Enc},\mathsf{Dec},\mathcal{Z}\}, D)=[\log D(x)+\log(1-D(\hat{x}))],
\end{align}
where $\|x-\hat{x} \|_{\text{precept}}$ is the perceptual reconstruction loss instead of $L_{2}$ loss, $\text{sg}[\cdot]$ denotes the stop-gradient operation, and $\|\text{sg}[v_{\textbf{q}}]-\mathsf{Enc}(x) \|_{2}^2$ is the commitment loss~\cite{van2017neural}. A patch-based discriminator $D$ is introduced for improving the quality of generated images by adversarial learning~\cite{goodfellow2014generative}. 

In this work, we use the pre-trained VQGAN weights for DAT directly, which are opened on GitHub\footnote{https://github.com/CompVis/taming-transformers}\footnote{https://github.com/CompVis/latent-diffusion}. VQGAN with $f=8$, $d=4$ and $K=16384$ is used for main experiments. We acknowledge the authors of~\cite{esser2021taming, rombach2021high}, whose works have greatly promoted our DAT.

\section{More Experimental Results}
\label{app: 2}
\subsection{Strategies Comparison on ResNet50}

Table~\ref{tab:strategy_compare_resnet50} presents the comparison results of DAT with other robust training strategies on ResNet50. Note that Advprop, Fast Advprop and Debiased use auxiliary BN while Pyramid AT and DAT not. Advprop achieves the best clean accuracy. The reason may lie in the auxiliary BN reduces the impact of adversarial examples on standard performance. On ResNet50, DAT is not always the best on the robustness benchmarks. By augmenting with the style transferred data, Debiased achieves best on IN-C, IN-Sketch and Stylized IN. On the contrary, such robust training method proposed for ViTs, e.g. Pyramid AT, is not incompatible with ResNet50 and obtain unsatisfactory results. However, our DAT is a general method which works for both CNNs and ViTs, and can be more practical.

%%%%%%%%%%%%%%%%%%%%%%%%%%%%%%%%%%%%%%%%%%%%%
\begin{table}[h]\centering
\scriptsize
\begin{tabular}{l|c|cc|ccc|ccc}
\toprule
Training & \multirow{ 2}{*}{ImageNet} & \multicolumn{2}{c|}{Adversarial Robustness} & \multicolumn{6}{c}{Out of Distribution Robustness} \\
 Strategies &  & FGSM & DamageNet & A & C$\downarrow$ & V2 & R & Sketch & Stylized \\
 \midrule
 Normal~\cite{steiner2021train} & 76.13 & 12.19 & 5.94 & 0 & 76.70 & 63.2 & 36.17 & 24.09 & 7.38 \\
 Advprop~\cite{xie2020adversarial} & \textbf{77.59} & 28.65 & \textbf{15.58} & 4.33 & 70.53 & \textbf{65.47} & 38.75 & 25.51 & 7.99 \\
 Fast Advprop~\cite{mei2021fast} & 76.6 & 17.33 & 7.45 & 2.19 & 73.31 & 64.24 & 38.17 & 25.03 & 8.3 \\
 Pyramid AT~\cite{herrmann2021pyramid} & 75.46 & 30.35 & 14.22 & 3.01 & 76.42 & 62.46 & 38.85 & 23.76 & 10.41 \\
 Debiased~\cite{li2020shape} & 76.91 & 20.4 & 6.66 & 3.51 & \textbf{67.55} & 65.04 & 40.8 & \textbf{28.42} & \textbf{17.4} \\
 \textbf{DAT (Ours)} & 76.52 & \textbf{30.66} & 14.42 & \textbf{4.38} & 74.16 & 65.02 & \textbf{41.9} & 27.27 & 10.8 \\
\bottomrule
 
\end{tabular}
\vspace{1em}
\caption{Comparison of DAT with other training strategies. We use ResNet50 as the base model.}
\label{tab:strategy_compare_resnet50}
\vspace{-1.5em}
\end{table}

\subsection{Comparing with Traditional Adversarial Training }

We compare our DAT with open-sourced adversarially robust models\footnote{https://github.com/microsoft/robust-models-transfer} in Table~\ref{tab:compare_adv_resnet50}. From the results, we can see a clear quantification of the benefit of our proposed discrete AT scheme compared with traditional AT. For clean performance, traditional AT plays a negative impact. However DAT can reduce the negative impact and achieve higher accuracy on validation set of ImageNet. It even surpasses the clean performance of normal training. The results also suggest AT with a very small $\epsilon$ can slightly benefit the generalization, e.g., with $\epsilon$=0.01 L2 AT, ImageNet-C mCE value from 76.70 drops to 75.33. But with the $\epsilon$ becoming larger, AT greatly damages the generalization, e.g. with $\epsilon$=5.0 L2 AT, ImageNet-C mCE value increases to 88.98. In contrast, our DAT achieves significant improvement on generalization compared with traditional AT.

\begin{table}[h]\centering
\scriptsize
    \begin{tabular}{l|c|c|ccc|ccc}
    \toprule
        Models & Train Cost & ImageNet & A & C↓ & V2 & R & Sketch & Stylized \\ \midrule
        Normal training, $\epsilon$=0 & 1× & 76.13 & 0.0 & 76.70 & 63.20 & 36.17 & 24.09 & 7.38 \\ \midrule
        L2-Robust, $\epsilon$=0.01 & 4× & 75.68 & 2.11 & 75.33 & 64.00 & 35.98 & 23.55 & 7.47 \\ 
        L2-Robust, $\epsilon$=0.03 & 4× & 75.76 & 2.17 & 75.36 & 63.66 & 36.18 & 23.98 & 8.18 \\ 
        L2-Robust, $\epsilon$=0.05 & 4× & 75.59 & 2.19 & 75.65 & 63.37 & 36.48 & 23.90 & 8.51 \\ 
        L2-Robust, $\epsilon$=0.1 & 4× & 74.78 & 2.13 & 75.42 & 62.64 & 36.90 & 23.85 & 9.18 \\ 
        L2-Robust, $\epsilon$=0.25 & 4× & 74.14 & 2.28 & 75.79 & 62.20 & 37.57 & 24.33 & 10.07 \\ 
        L2-Robust, $\epsilon$=0.5 & 4× & 73.16 & 2.19 & 75.91 & 60.48 & 38.03 & 23.49 & 10.99 \\ 
        L2-Robust, $\epsilon$=1.0 & 4× & 70.43 & 2.19 & 78.36 & 57.36 & 38.21 & 22.63 & 11.07 \\ 
        L2-Robust, $\epsilon$=3.0 & 4× & 62.83 & 1.97 & 83.84 & 49.45 & 36.48 & 20.40 & 10.48 \\ 
        L2-Robust, $\epsilon$=5.0 & 4× & 56.13 & 1.71 & 88.98 & 43.04 & 32.75 & 16.82 & 9.13 \\ \midrule
        Linf-Robust, $\epsilon$=0.5/255 & 4× & 73.73 & 2.35 & 76.86 & 61.88 & 38.54 & 23.79 & 10.94 \\ 
        Linf-Robust, $\epsilon$=1.0/255 & 4× & 72.05 & 2.53 & 78.34 & 59.60 & 40.13 & 23.70 & 12.10 \\ 
        Linf-Robust, $\epsilon$=2.0/255 & 4× & 69.10 & 2.52 & 80.09 & 56.64 & 38.65 & 22.14 & 12.36 \\ 
        Linf-Robust, $\epsilon$=4.0/255 & 4× & 63.86 & 2.25 & 85.14 & 51.39 & 38.25 & 20.94 & 11.70 \\ 
        Linf-Robust, $\epsilon$=8.0/255 & 4× & 54.53 & 2.12 & 91.59 & 42.16 & 34.40 & 18.10 & 9.58 \\ \midrule
        DAT (Ours) & 3.5× & 76.52 & 4.38 & 74.16 & 65.02 & 41.90 & 27.27 & 10.8 \\ \bottomrule
    \end{tabular}
    \vspace{1em}
    \caption{Comparison of our DAT with adversarial training models.}
    \vspace{-1.5em}
\label{tab:compare_adv_resnet50}
\end{table}

%%%%%%%%%%%%%%%%%%%%%%%%%%%%%%%%%%%%%%%%%%%%%

\subsection{DAT with Different Perturbations}
Table~\ref{tab:modify_tokens} counts for the percentage of the modified visual words with different magnitude $\alpha$. We show only 3.8\% of the visual words are changed when we set $\alpha$ as 0.1. This proportion does not have negative impact on the trained models, but even somewhat benefits the clean accuracy. With larger proportion of visual words being adversarially altered, the standerd performance and generalization are getting worse. We also experiment DAT with random perturbations. To introducing the randomness, we selects 3.8\% visual words and replaces them with other words. We find this operation can slightly improvement the generalization of learned representation, but it still cannot achieve the comparable effect as our DAT. It suggests adversarially altering the visual words is a better way.

\begin{table}[h]\centering
\scriptsize
\begin{tabular}{l|cc|c|cc|ccc|ccc}
\toprule
 & & Modified & \multirow{ 2}{*}{ImageNet} & \multicolumn{2}{c|}{Adversarial Robustness} & \multicolumn{6}{c}{Out of Distribution Robustness} \\
 Types & $\alpha$ & Codes &  & FGSM & DamageNet & A & C$\downarrow$ & V2 & R & Sketch & Stylized \\
 \midrule
 Random & - & 3.8\% & 76.47 & 29.01 & 10.7 & 3.00 & 74.71 & 64.75 & 40.19 & 26.17 & 9.89 \\
 Adv. & 0.0 & 0.0\% & 76.38 & 23.94 & 9.12 & 3.2 & 76.31 & 64.71 & 38.41 & 24.62 & 8.77 \\
 Adv. & 0.1 & 3.8\% & 76.52 & 30.66 & 14.42 & 4.38 & 74.16 & 65.02 & 41.9 & 27.27 & 10.8 \\
 Adv. & 0.2 & 7\% & 75.93 & 34.47 & 15.21 & 3.11 & 75.09 & 64.38 & 40.27 & 26.33 & 10.14 \\
 Adv. & 0.4 & 13\% & 74.28 & 36.2 & 17.76 & 1.96 & 77.25 & 62.5 & 38.75 & 24.31 & 8.61 \\
\bottomrule
 
\end{tabular}
\vspace{1em}
\caption{DAT with different perturbations. We use ResNet50 as the base model.}
\label{tab:modify_tokens}
\vspace{-1.5em}
\end{table}

\subsection{The Effect of DAT on Corruptions in ImageNet-C }

To analyse the effect of DAT on each image corruption in ImageNet-C, we report the detailed results in Table~\ref{tab:imagenet-c}. For ResNet50, we find DAT reduces the accuracy on images with contrast and fog corruptions. It demonstrates that ResNet50 trained by DAT can be sensitive to the image lack of the hierarchy. However, for ViT, DAT can improve the performance on all corruptions. It suggests DAT works more efficiently on transformer-based vision models.

%%%%%%%%%%%%%%%%%%%%%%%%%%%%%%%%%%%%%%%%%%%%%
\begin{table*}[h]
\centering
% \captionsetup{font={footnotesize}}
\label{tab:robustness_imc}
\vspace{2mm}
% \vspace{-4pt}
% \vspace{-3mm}
\setlength{\tabcolsep}{0.5mm}
% \vspace{-5pt}
\resizebox{\textwidth}{!}{

\addtolength{\tabcolsep}{2pt}
\begin{tabular}{l|c|cccc|ccc|cccc|cccc}
% \toprule
\multirow{2}{*}{Model} & \multirow{2}{*}{Average } & \multicolumn{4}{c|}{Blur} & \multicolumn{3}{c|}{Noise} & \multicolumn{4}{c|}{Digital} & \multicolumn{4}{c}{Weather} \\
\cline{3-17}
& & Motion & Defoc & Glass & \multicolumn{1}{c|}{Zoom} & Gauss & Impul & \multicolumn{1}{c|}{Shot} & Contr & Elast & JPEG & \multicolumn{1}{l|}{Pixel} & \multicolumn{1}{l}{Bright.} & \multicolumn{1}{l}{Snow} & \multicolumn{1}{l}{Fog} & \multicolumn{1}{l}{Frost} \\
\toprule
\multicolumn{1}{l|}{ResNet50} & \multicolumn{1}{c|}{39.2} & 38.7 & 38.8 & 26.8 & \multicolumn{1}{c|}{36.2} & 29.2 & 23.8 & \multicolumn{1}{c|}{27.0} & 39.1 & 45.3 & 53.4 & \multicolumn{1}{c|}{44.8} & 68.0 & 32.5 & 45.8 & 38.1 \\

\multicolumn{1}{l|}{+DAT (Ours)} & \multicolumn{1}{c|}{41.1} & 38.3 & 37.2 & 33.7 & \multicolumn{1}{c|}{37.9} & 33.0 & 28.1 & \multicolumn{1}{c|}{31.1} & 36.5 & 50.0 & 59.0 & \multicolumn{1}{c|}{45.6} & 69.0 & 34.2 & 41.5 & 41.0 \\

\midrule
\multicolumn{1}{l|}{ViT} & \multicolumn{1}{c|}{57.2} & 54.2 & 47.9 & 43.0 & \multicolumn{1}{c|}{41.6} & 61.9 & 58.4 & \multicolumn{1}{c|}{58.3} & 60.2 & 58.0 & 61.4 & \multicolumn{1}{c|}{65.9} & 74.8 & 52.1 & 61.6 & 59.5 \\
\multicolumn{1}{l|}{+DAT (Ours)} & \multicolumn{1}{c|}{65.2} & 58.4 & 55.1 & 49.8 & \multicolumn{1}{c|}{50.8} & 71.3 & 70.3 & \multicolumn{1}{c|}{70.5} & 71.7 & 63.1 & 69.1 & \multicolumn{1}{c|}{67.7} & 78.2 & 64.0 & 69.6 & 68.2 \\

\end{tabular}
}
\caption{Detailed results of DAT on each image corruption in ImageNet-C.}
\label{tab:imagenet-c}
% \vspace{-3mm}
\end{table*}
%%%%%%%%%%%%%%%%%%%%%%%%%%%%%%%%%%%%%%%%%%%%%

\subsection{Training Budget for DAT}
DAT is experimented on 32 2080Ti GPUs. We compare the training cost with other robust training strategies in same setting. The results is shown in Table~\ref{tab:training_cost}. DAT only needs one attack step to generate discrete adversarial examples, which makes it less expensive than standard adversarial training. However, DAT still requires 3.5$\times$ training budget than normal training. We believe that reducing the cost of DAT is necessary, which will be remained as the future work. 

\begin{table}[h]\centering
\scriptsize
\begin{tabular}{l|c|c}
\toprule
Training Strategies & Attack Steps & Training Budget \\
 \midrule
 Normal & 0 & 1$\times$ \\
 Adversarial Training & 10 & 11$\times$ \\
 Advprop & 5 & 7$\times$ \\
 Advprop & 1 & 3$\times$ \\
 DAT (Ours) & 1 & 3.5$\times$ \\
\bottomrule
\end{tabular}
\vspace{1em}
\caption{Comparison of the training costs.}
\label{tab:training_cost}
\vspace{-1.5em}
\end{table} 

\subsection{Inference with the Discretizer $\mathcal{Q}$}

In this work, we delete the discretizer $\mathcal{Q}$ at inference time. However, there is another option that remaining the discretizer for test inputs. To study the effect of this alternative, we report some results in Table~\ref{tab:inference}. Although inference with $\mathcal{Q}$ brings improvement on adversarial robustness, it meanwhile harms the standard performance and generalization. The inference cost also increases by the additional computation on $\mathcal{Q}$. Therefore, such alternative cannot yield the best trade-off on speed and performance, which is not adopted by our DAT.

\begin{table}[h]\centering
\scriptsize
\begin{tabular}{l|c|cc|ccc|ccc}
\toprule
 & \multirow{ 2}{*}{ImageNet} & \multicolumn{2}{c|}{Adversarial Robustness} & \multicolumn{6}{c}{Out of Distribution Robustness} \\
 Methods &  & FGSM & DamageNet & A & C$\downarrow$ & V2 & R & Sketch & Stylized \\
 \midrule
 ResNet50 + DAT (w/o $\mathcal{Q}$) & 76.52 & 30.66 & 14.42 & 4.38 & 74.16 & 65.02 & 41.90 & 27.27 & 10.8 \\
 ResNet50 + DAT (w/ $\mathcal{Q}$) & 74.8 & 55.4 & 20.44 & 4.13 & 76.06 & 63.04 & 39.67 & 25.64 & 10.16 \\
 ViT + DAT (w/o $\mathcal{Q}$) & 81.46 & 51.82 & 45.70 & 30.15 & 44.65 & 70.83 & 47.34 & 34.77 & 23.13 \\
 ViT + DAT (w/ $\mathcal{Q}$) & 80.12 & 61.65 & 50.4 & 22.44 & 48.43 & 68.59 & 47.2 & 34.41 & 21.83 \\
\bottomrule
\end{tabular}
\vspace{1em}
\caption{The ablation on the discretizer $\mathcal{Q}$ at inference time.}
\label{tab:inference}
\vspace{-1.5em}
\end{table} 

\subsection{Is it necessary for bounding $\delta$?}
Traditional adversarial attacks always add constraints on perturbations. While in this work, there are no restrictions on $\delta$. To explore if bounding $\delta$ is necessary in DAT, we add $l_{\infty}$ bound on $\delta$ with different $\epsilon$ and re-run the DAT. The result is shown in Table~\ref{tab:bound_delta}, DAT achieves best performance when $\delta$ is not bounded. The worst result is appeared when $\delta$ is bounded with $\epsilon=1/255$. With larger $\epsilon$, the results become better. Therefore, it seems bounding $\delta$ in our DAT is not necessary, and even plays negative affect on the overall performance.

\begin{table}[h]\centering
\scriptsize
\begin{tabular}{l|c|cc|ccc|ccc}
\toprule
 & \multirow{ 2}{*}{ImageNet} & \multicolumn{2}{c|}{Adversarial Robustness} & \multicolumn{6}{c}{Out of Distribution Robustness} \\
 $\epsilon$ of $l_{\infty}$ &  & FGSM & DamageNet & A & C$\downarrow$ & V2 & R & Sketch & Stylized \\
 \midrule
 1/255 & 76.10 & 29.41 & 12.00 & 3.53 & 75.53 & 64.11 & 39.05 & 25.04 & 8.69 \\
 2/255 & 76.16 & 29.75 & 13.24 & 3.75 & 74.87 & 64.32 & 40.38 & 25.53 & 9.31 \\
 4/255 & 76.47 & 31.43 & 14.25 & 4.31 & 74.12 & 65.07 & 41.68 & 26.99 & 10.62 \\
 $\infty$ (Ours) & 76.52 & 30.66 & 14.42 & 4.38 & 74.16 & 65.02 & 41.90 & 27.27 & 10.8 \\

\bottomrule
\end{tabular}
\vspace{1em}
\caption{The trained model on DAT when $\delta$ is bounded by different $\epsilon$.}
\label{tab:bound_delta}
\vspace{-1.5em}
\end{table} 

\subsection{DAT for Domain Generalization}
In addition to training on large-scale ImageNet, we evaluate our DAT on Domain Generalization (DG) tasks. DG task is more challenging since it needs the learned model to transfer between multiple domains, using only small amount of the data. We adopt PACS dataset, which consists of four domains, namely Photo, Art Painting, Cartoon and Sketch. Each domain contains seven categories. For each trial, we train on 3 domains for generalizing to remaining unseen domain. To keep the setting consistent with previous works, we use AlexNet as the backbone. The results are shown in below Table~\ref{tab:dat_on_dg}. We only compare with previously adversarial augmentation based DG methods. DAT can also achieve better domain generalization performance on PACS dataset. It has slight drop on domain of photo, but improves the transferability on other three domains.

\begin{table}[h]\centering
\scriptsize
\begin{tabular}{c|c|c|c}
\toprule
 & ADA & MD-ADA & DAT (Ours) \\
 \midrule
 Art Painting & 64.3 & 67.1 & \textbf{67.3}  \\
 Cartoon & 69.8 & 69.9 & \textbf{71.3} \\
 Photo & 85.1 & \textbf{88.6} & 87.8 \\
 Sketch & 60.4 & 63.0 & \textbf{64.1} \\
 \midrule
 Average & 69.9 & 72.2 & \textbf{72.6} \\
\bottomrule
\end{tabular}
\vspace{1em}
\caption{Classification accuracy (\%) of our DAT on PACS dataset in comparison with the
previously adversarial augmentation based DG methods. }
\label{tab:dat_on_dg}
\vspace{-1.5em}
\end{table}

\section{Implementation of the Stronger Attacks in Ablation Experiments}
\label{app: 3}
For AutoAttack, we attack a subset of ImageNet provided in RobustBench~\footnote{https://github.com/RobustBench/robustbench}, which consists of 5000 images. Three perturbations bounded with $l_{\infty}$-norm of $1/255$, $0.5/255$ and $0.25/255$ are adopted. For AdvDrop, the test dataset is 1000 random sampled images on ImageNet, which is provided in the official implementaton~\footnote{https://github.com/RjDuan/AdvDrop}. The bound of the quantization table is a key factor in AdvDrop which controls the attack strength. We use three bounds with 10, 15 and 20 to regularize the quantization table. For PerC-Adversarial, we use the Perceptual Color distance Alternating Loss (PerC-AL) method to generate adversarial examples. PerC-Adversarial uses the test data of Defense Against Adversarial Attack Challenge in NeurIPS 2017. We change the attack strength by three different attack iterations: 20, 40 and 60.

%%%%%%%%%%%%%%%%%%%%%%%%%%%%%%%%%%%%%%%%%%%%%
\begin{figure}
  \centering
  \includegraphics[width=1.0\linewidth]{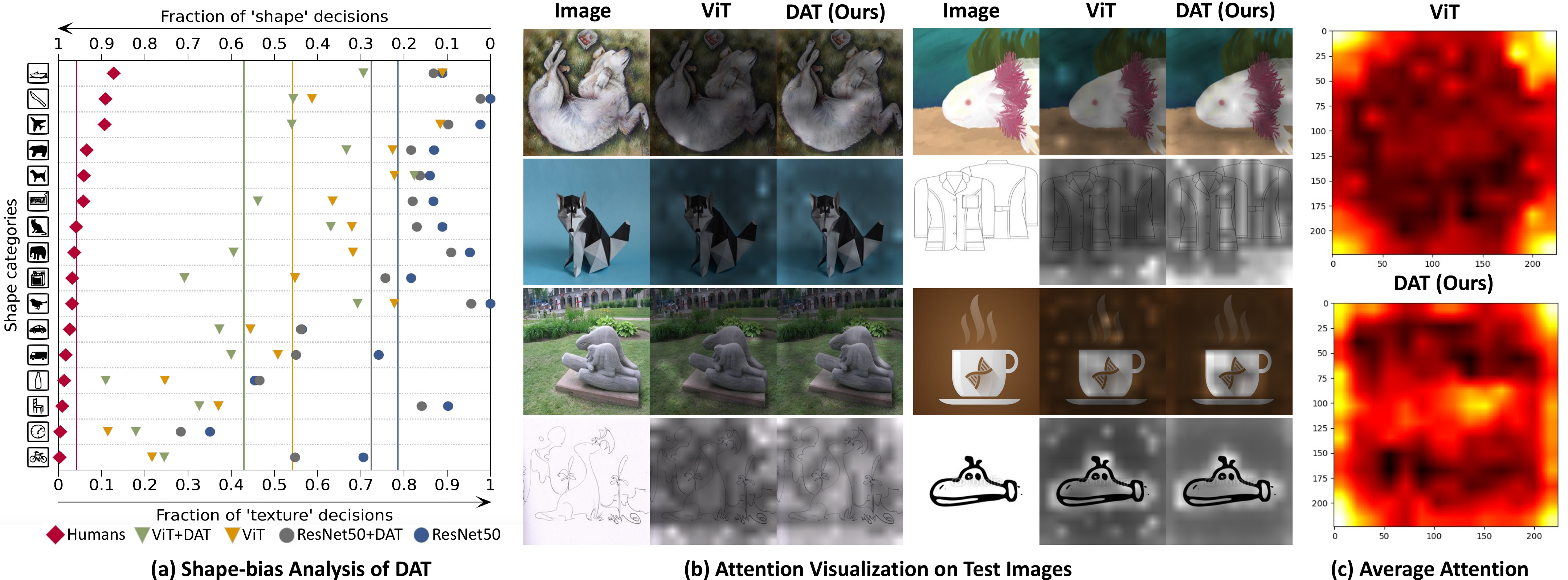}
  \caption{(a) The fraction of correct shape-based decisions of models w/ and w/o DAT. (b) Visualized attention on test images of ImageNet-R. (c) The heat map of averaged attention with ViT and ViT trained by our DAT.}
  \label{fig:attn_vis}
  \vspace{-1em}
\end{figure}
%%%%%%%%%%%%%%%%%%%%%%%%%%%%%%%%%%%%%%%%%%%%%

%%%%%%%%%%%%%%%%%%%%%%%%%%%%%%%%%%%%%%%%%%%%%
\begin{figure}
  \centering
  \includegraphics[width=1.\linewidth]{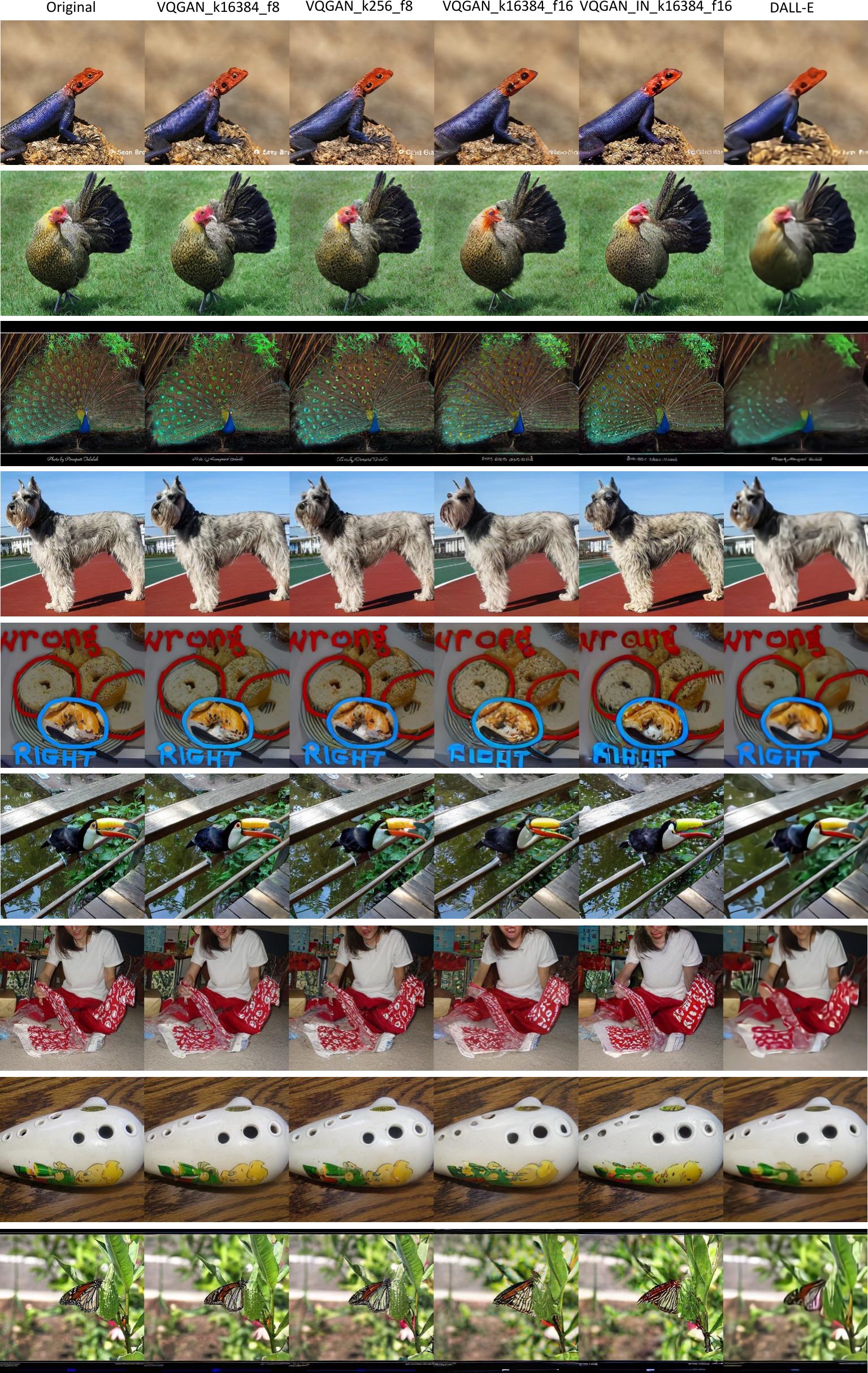}
  \caption{The visualization of the discrete reconstruction $\hat{x}$ based on different discretizers.}
  \label{fig:rec_vis}
  \vspace{-1em}
\end{figure}
%%%%%%%%%%%%%%%%%%%%%%%%%%%%%%%%%%%%%%%%%%%%%

%%%%%%%%%%%%%%%%%%%%%%%%%%%%%%%%%%%%%%%%%%%%%
\begin{figure}
  \centering
  \includegraphics[width=1.\linewidth]{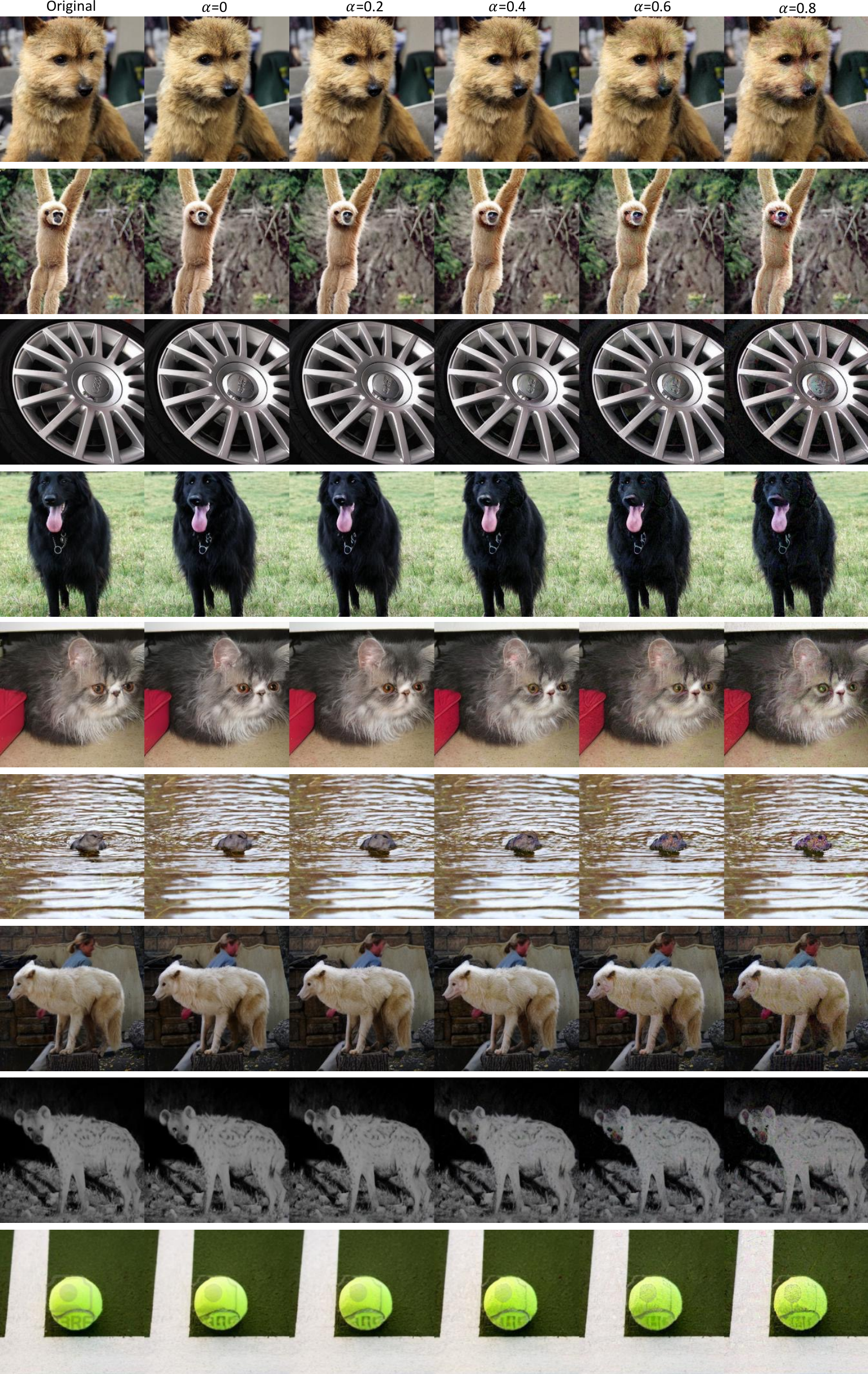}
  \caption{Training example visualization of DAT with different $\alpha$.}
  \label{fig:adv_vis}
  \vspace{-1em}
\end{figure}
%%%%%%%%%%%%%%%%%%%%%%%%%%%%%%%%%%%%%%%%%%%%%

\section{Visualization}
\label{app: 4}
\subsection{Attention Visualization}
We visualize the attention of ViT trained by DAT in Figure~\ref{fig:attn_vis}(b). For the object with unusual renditions in ImageNet-R, ViT cannot attend to semantically relevant image regions. While our DAT can locate the attention to the central object more related for the classification. This phenomenon is also reflected by the statistical average of the attention in Figure~\ref{fig:attn_vis}(c). By randomly sampling 1000 images in ImageNet-R and averaging the attention maps, we find the attentions of our DAT are more global. Compared with ViT which puts much attentions on the corners of the image found by prior work~\cite{dosovitskiy2020image}, our DAT additionally attends the central regions in where the classified object is often located.

\subsection{Shape-bias Analysis}
We conduct an analysis based on shape-bias, which represents the fraction of correct decisions based on object shape. The result is shown in Figure~\ref{fig:attn_vis}(a). The averaged scores on 16 categories is denoted by the colored vertical line. We compare decisions with Humans, ResNet50 w/ and w/o DAT, ViT w/ and w/o DAT. Human decisions are highly based on shape, which achieve the best average fraction of 0.96 to correctly recognize an image by shape. By comparison, ViT and ResNet50 still have large gap with humans on the ability of learning shape features. In this work, we find our DAT can help for improving the fraction of shape-based decisions of models. It suggests DAT regularizes the models to learn texture independent shape features, and behave more like a human.

\subsection{Comparison of the Reconstruction Quality}
We compare the reconstruction quality of different discretizers $\mathcal{Q}$ in Figure~\ref{fig:rec_vis}. VQGAN with $k=16384$ and $f=8$ can retain the most of the image details, which has the least impact on the classification models when used for training. With the growing of the downsampling factor $f$, some fine-grained attributes of the objects can be changed. For example, in third row the spotted texture on the peacock's tail is partly lost after reconstruction. Compared with VQGAN models, DALL-E blurs the image to a greater extent, yielding a low quality reconstructed image. Accordingly, DAT which uses DALL-E for image discretization performs worst on clean accuracy in Table~\ref{tab:codebooktype}. It shows that the reconstruction quality is indeed proportional to the standard performance in DAT.  

\subsection{Visualization of Discrete Adversarial Examples}
We visualize the discrete adversarial examples which is generated by DAT for training in Figure~\ref{fig:adv_vis}. With the growing of $\alpha$, the alteration on images become larger. However, different from traditional adversarial example which essentially adds global high-frequency noise on images, we find DAT modifies the properties of the local part of the object. For example, in the fifth row of Figure~\ref{fig:adv_vis}, the discrete adversarial example is changing the eye color of the cat. Such modification is large but imperceptible and semantic-preserving. There are also some failure case where DAT changes the semantics. In sixth row of Figure~\ref{fig:adv_vis}, after discrete reconstruction, the otter in image looks more like a dog. We believe that such cases are only a minority, and will not affect the overall performance of DAT.

\subsection{Visualization of the Straight-Through Gradients and Backward Gradients in DAT }

\begin{figure}
  \centering
  \includegraphics[width=1.\linewidth]{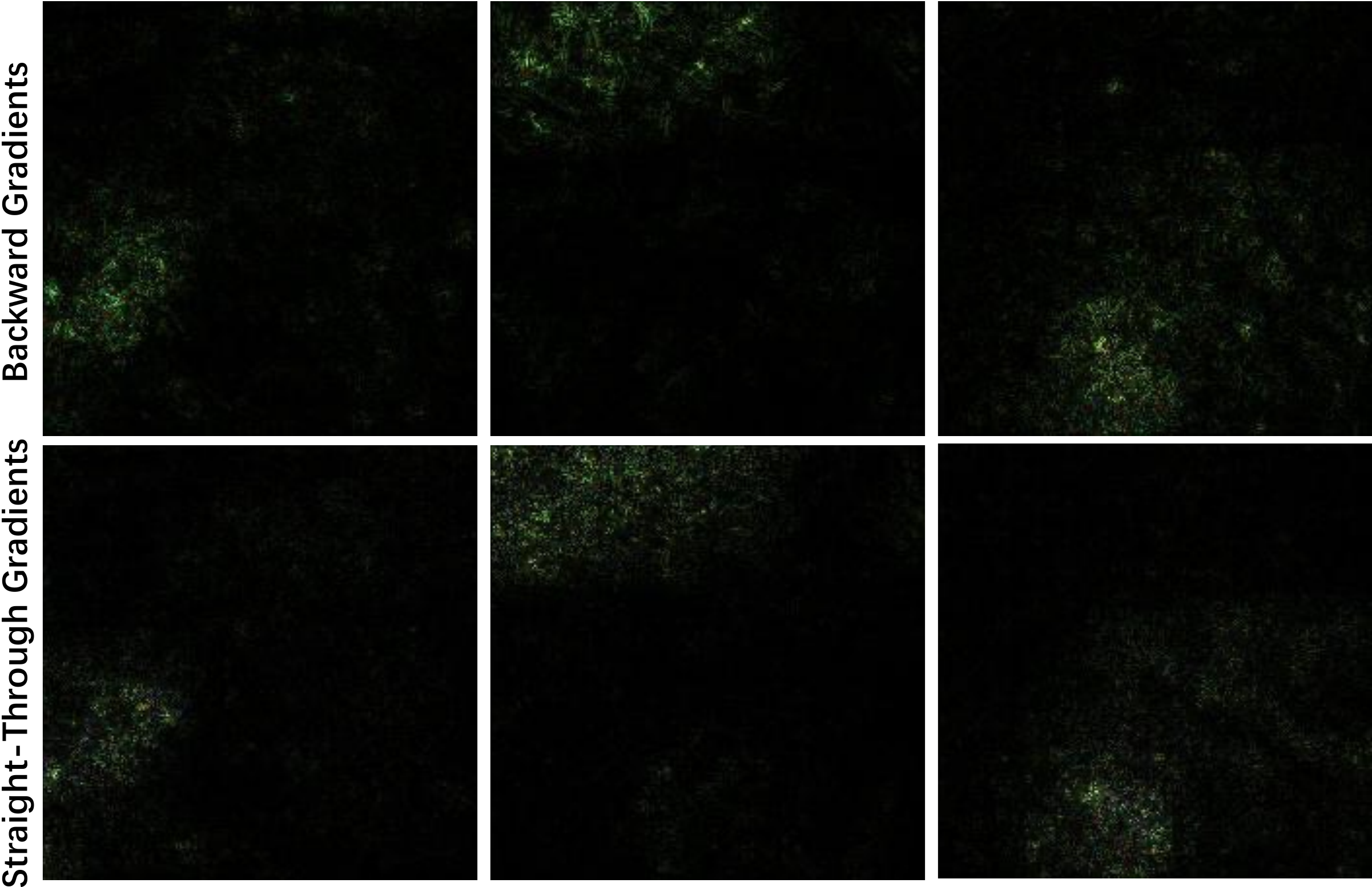}
  \caption{Visualization of the straight-through gradients and backward gradients. }
  \label{fig:grad_compare}
  \vspace{-1em}
\end{figure}

We visualize the estimated straight-through gradients in Eq~\ref{eq:final} and the directly backward gradients in Eq~\ref{eq:expand}. As shown in Figure~\ref{fig:grad_compare}, the approximated gradient by straight-through estimator does accurately estimate the ground-truth gradients in Eq~\ref{eq:expand}. It reflects the rationality of the proposed efficient straight-through gradient in DAT.

%%%%%%%%%%%%%%%%%%%%%%%%%%%%%%%%%%%%%%%%%%%%%
%%%%%%%%%%%%%%%%%%%%%%%%%%%%%%%%%%%%%%%%%%%%%

\end{document}